\documentclass[11pt]{article}

\usepackage[final]{acl}

\usepackage{times}
\usepackage{latexsym}

\usepackage[T1]{fontenc}

\usepackage[utf8]{inputenc}

\usepackage{microtype}

\usepackage{inconsolata}

\usepackage{graphicx}
\usepackage{multirow}
\usepackage{booktabs}
\usepackage{amsmath}
\usepackage{cleveref}
\usepackage{amssymb}
\usepackage[table]{xcolor}
\usepackage{colortbl}

\definecolor{lightgray}{gray}{0.92}

\newcommand{\methodname}{\textsc{Otter}}

\title{What Matters When Building Universal Multilingual Named Entity Recognition Models?}

\author{
    Jonas Golde \hspace{1em} Patrick Haller  \hspace{1em} Alan Akbik \\
    Humboldt Universität zu Berlin \\
    \texttt{jonas.max.golde.1@hu-berlin.de}
}

\begin{document}
\maketitle
\begin{abstract}
Recent progress in universal multilingual named entity recognition (NER) has been driven by multilingual transformer models, task-specific architectures, custom loss functions, and large-scale training datasets. However, despite substantial prior work, we find that many design decisions for such models are made without systematic justification, with individual components evaluated only in combination rather than in isolation. We argue that this impedes progress in the field by making it difficult to identify which choices improve multilingual generalization. In this work, we conduct an extensive empirical evaluation on transformer backbones, architectures, training objectives, data composition, and threshold selection for zero-shot, multilingual NER models. Building on these findings, we present \methodname{}, a small encoder model that achieves consistent improvements over strong multilingual NER baselines, outperforming similarly sized models by 5.7 percentage points in F1. Further, it remains within 1.4 points of Qwen3-32B and 5.4 points below Gemma3-27B, while being roughly $90\times$ smaller and one to two orders of magnitude faster at inference.
\end{abstract}

\section{Introduction}

\begin{figure}[!ht]
    \includegraphics[width=\linewidth]{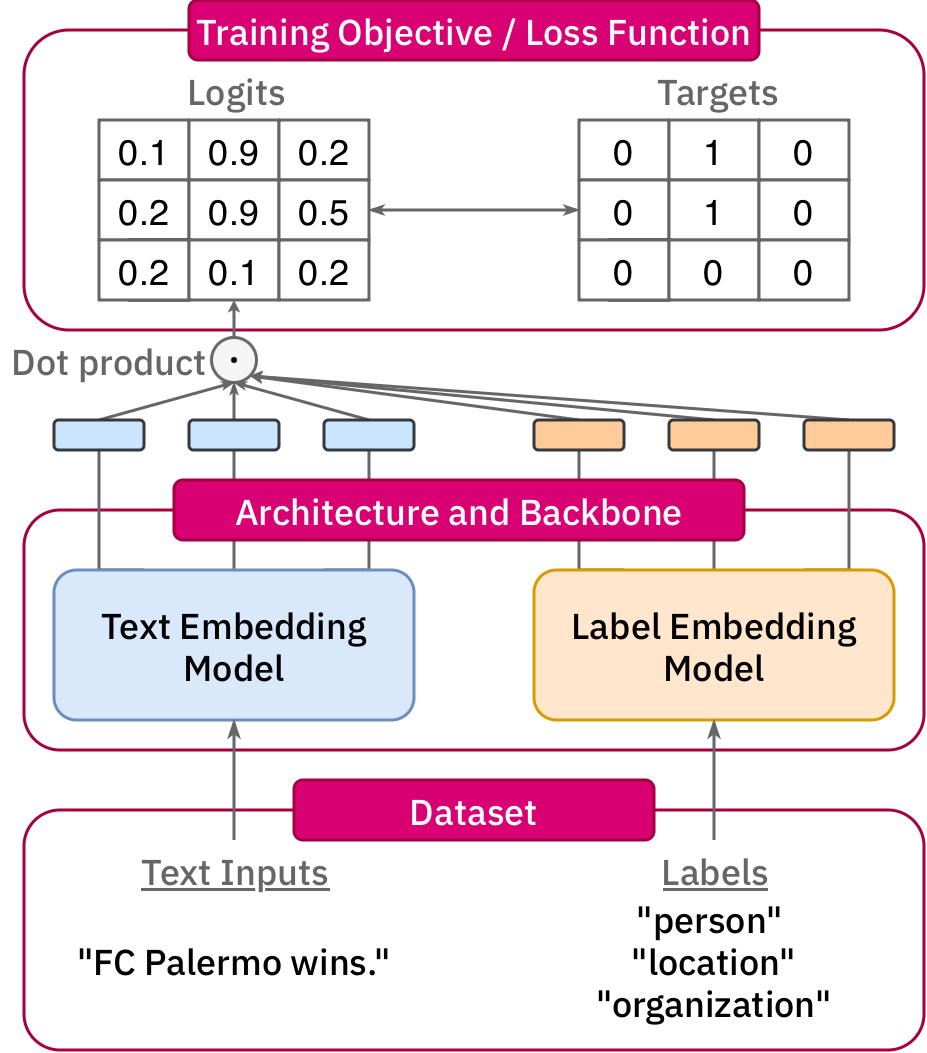}
    \caption{Overview of the design choices in universal multilingual NER systems: the dataset, backbone, architecture, and training objective. Prior work typically fixes a particular combination of these choices and studies only their joint effect. In this work, we vary each dimension in a controlled setting to isolate its impact on performance and efficiency.}
    \label{fig:overview}
\end{figure}

Multilingual named entity recognition (NER) is a token-level information extraction task \citep{lample-etal-2016-neural,akbik-etal-2018-contextual} used in applications such as knowledge graph construction \citep{zhong2023comprehensive,arsenyan-etal-2024-large}, invoice parsing \citep{perot-etal-2024-lmdx} or web search \citep{shachar-etal-2025-ner}. While large language models (LLMs) \citep{grattafiori2024llama3herdmodels,yang2025qwen3technicalreport,deepseekai2025deepseekv3technicalreport,gemmateam2025gemma3technicalreport} are increasingly capable and improve performance on multilingual NER, their billion parameters come at a high computational cost. Instead, recent work uses such models as effective teachers for training smaller, more efficient encoder-only models, as done in GLiNER \citep{zaratiana-etal-2024-gliner,zaratiana-etal-2025-gliner2} or NuNER \citep{bogdanov-etal-2024-nuner}.

\paragraph{Design choices are rarely justified.}~However, despite this progress, many design choices in related work are only weakly justified or briefly discussed. Because such choices vary widely across systems, it is hard to conclude which ones actually contribute to better generalization across languages and domains. For instance, \citet{zaratiana-etal-2024-gliner,huang-etal-2022-copner} use a cross-encoder architecture, while \citet{zhang2023optimizing,bogdanov-etal-2024-nuner,golde-etal-2024-large} use a bi-encoder. Likewise, \texttt{gliner-multi-v2.1}\footnote{\url{https://huggingface.co/urchade/gliner_multi-v2.1}} uses mDeBERTa \citep{he2023debertav3improvingdebertausing} as its backbone, whereas \texttt{gliner-x-base}\footnote{\url{https://huggingface.co/knowledgator/gliner-x-base}} uses mT5 \citep{xue-etal-2021-mt5}. To the best of our knowledge, these choices have never been systematically ablated, so their trade-offs in compute, data efficiency, and performance remain poorly understood. In this work, we examine these design dimensions and measure their effects on performance and efficiency.

\paragraph{Identifying the design dimensions.}~We investigate five core dimensions arising from prior work: \textit{(i)} the choice between cross-encoder and bi-encoder architectures, \textit{(ii)} the transformer backbone, \textit{(iii)} the training objective, \textit{(iv)} the composition of the training dataset, and \textit{(v)} the selection of language-specific thresholds for zero-shot NER. \Cref{fig:overview} illustrates these dimensions. In this paper, we empirically evaluate these dimensions within a controlled experimental setting to isolate their individual and joint effects.

\paragraph{Contributions.}~Building on these insights, we train \methodname{}, a universal multilingual NER model optimized for the long tail of languages, which we evaluate on more than 150 distinct languages. \methodname{} achieves strong zero-shot performance across seven benchmarks, consistently outperforming existing models of comparable size while running one to two orders of magnitude faster than comparable autoregressive baselines (\Cref{sec:efficiency_appendix}). We summarize our contributions as follows:
\begin{itemize}
    \item We empirically investigate the design choices in universal multilingual NER, isolating the effects of architecture, transformer backbone, loss function, training data and language-specific thresholding within a staged, controlled comparison.
    \item We generally find that a simple loss function and large-scale multilingual training are sufficient to obtain strong performance. Further, we support our findings with several ablations, such as score calibration, the impact of subword fragmentation, and candidate-span imbalance.
    \item Based on our findings, we present \methodname{}, a multilingual NER model supporting over 100 languages, and evaluate it across seven benchmarks, achieving state-of-the-art zero-shot performance on many of them.
    \item We release the model checkpoints as well as the training and evaluation code to support reproducibility and further research using our model.\footnote{\url{https://github.com/whoisjones/otter}}
\end{itemize}

\section{Methodology}

\subsection{Dimension 1: Architecture and Backbone}~At a high level, there are two main approaches for combining text inputs $\mathbf{x} = \mathrm{x}_1, \ldots, \mathrm{x}_n$ and label inputs $\mathbf{y} = \mathrm{y}_1, \ldots, \mathrm{y}_n$. The cross-encoder approach concatenates text and label descriptions and processes them jointly through a single transformer, allowing text tokens and label tokens to cross-attend to each other (\Cref{eq:ce-encode}). This paradigm is used, for example, in GLiNER \citep{zaratiana-etal-2024-gliner}. In contrast, bi-encoders process text and label inputs separately using two transformer encoders (\Cref{eq:bi-encode}), as in Binder \citep{zhang2023optimizing}, and combine their representations only after encoding.
\begin{gather}
\mathbf{h}^{\mathrm{x}}, \mathbf{h}^{\mathrm{y}}
= f_{\operatorname{cross}}(\mathbf{x}, \mathbf{y}),
\label{eq:ce-encode} \\
\mathbf{h}^{\mathrm{x}}
= f_{\operatorname{bi}}^{text}(\mathbf{x}), \qquad
\mathbf{h}^{\mathrm{y}} = \operatorname{Pool}(f_{\operatorname{bi}}^{labels}(\mathrm{y})) .
\label{eq:bi-encode}
\end{gather}

where $\operatorname{Pool}(\cdot)$ maps the token embeddings of a label description to a single vector. For the cross-encoder setup, we follow the idea of GLiNER and introduce an additional \texttt{[LABEL]} token to obtain single-vector label representations. The underlying encoders are typically initialized from pretrained transformer models.

Given hidden representations $\mathbf{h}^{\mathrm{x}} \in \mathbb{R}^{|\mathrm{x}| \times d}$ and $\mathbf{h}^{\mathrm{y}} \in \mathbb{R}^{d}$, we project text and label representations using a two-layer MLP to obtain span-start, span-end, and label embeddings (\Cref{eq:s-proj,eq:e-proj,eq:label-proj}). We then concatenate the corresponding start and end representations with a span-width embedding to obtain a span representation $\mathbf{k}_{i,j}$ for each span $(i,j)$ (\Cref{eq:span-score}).
\begin{gather}
\mathbf{s}
= \operatorname{MLP}_{\operatorname{start}}(\mathbf{h}^{\mathrm{x}}),
\label{eq:s-proj}\\
\mathbf{e}
= \operatorname{MLP}_{\operatorname{end}}(\mathbf{h}^{\mathrm{x}}),
\label{eq:e-proj}\\
\mathbf{q}
= \operatorname{MLP}_{\operatorname{label}}(\mathbf{h}^{\mathrm{y}}),
\label{eq:label-proj}\\
\mathbf{k}_{i,j}
= \mathrm{MLP}_{\operatorname{span}} \big(
\mathbf{s}_i \oplus \mathbf{e}_j \oplus \mathbf{d}(j-i)
\big).
\label{eq:span-score}
\end{gather}

We then use this span representation $k_{i,j}$ to compute label-specific logits using label representations $\mathbf{q}$ (\Cref{eq:span-logits}).
\begin{gather}
\ell_{i,j}
= \mathbf{k}_{i,j}^{\top}\mathbf{q}
\label{eq:span-logits}
\end{gather}

This formulation extends naturally to $n$ labels by computing $\ell_{i,j,n} = \mathbf{k}_{i,j}^{\top}\mathbf{q}_{n}$ for every $\mathbf{q}_{n} \in \mathbf{Q}$. The underlying encoders are usually initialized from pretrained transformer models and trained using in-batch negatives. We note that more advanced negative mining strategies exist that go beyond negatives occurring within the batch, which we do not explore in this work.

\subsection{Dimension 2: Fine-Tuning Datasets}~
In parallel, several large-scale NER datasets have been released that use large language models to annotate unlabeled data. These datasets typically have large label sets following a long-tail distribution, such as PileNER \citep{zhou2024universalnertargeteddistillationlarge}, NuNER \citep{bogdanov-etal-2024-nuner} or LitSet \citep{golde-etal-2024-large}. When we train a smaller model on these datasets, the resulting supervision relies on noisy label descriptions rather than on a fixed set of classes. Since these label sets contain many synonymous and hierarchically related labels of varying granularity, the model learns to interpret the semantics of the label descriptions themselves, which is what allows the model to generalize to arbitrary, unseen labels at inference time. In our experiments, we use datasets that vary substantially in language coverage, ranging from English-only to 91 languages. This lets us investigate whether a multilingual transformer trained only on English is sufficient for cross-lingual generalization, or whether fine-tuning on multilingual data yields additional gains.

\paragraph{Multilingual tokenization.}~Because tokenization produces spans of different lengths across languages (e.g., shorter ones for English and longer ones for Chinese), the ratio of positive to negative span candidates varies by language, which affects the calibration of the decision threshold (cf. \Cref{sec:pre_tokenized_text_impact_appendix}). We deliberately use a single global threshold across all languages rather than tuning one per language, as the latter would require labeled data in each target language.

\subsection{Dimension 3: Loss Functions}~
Given the span-label scores $\ell_{i,j,n}$, the loss function determines how errors are converted into a training signal. In our setting, it must handle the severe class imbalance in span-based NER, where most span–label pairs are negatives. Existing approaches use different loss functions: Binder uses a contrastive loss to align span and label representations in a shared embedding space, whereas GLiNER applies a binary cross-entropy loss. In this work, we explore a range of loss functions that all aim to improve generalization under imbalanced positive-to-negative annotations: binary cross-entropy (BCE), BCE with positive weighting, focal loss \citep{lin2018focallossdenseobject} with varying $\alpha$ and $\gamma$, contrastive loss, and dice loss \citep{li-etal-2020-dice}.

\subsection{Modeling Without Word Segmentation}
\label{sec:no_word_seg}~
Classical NER relies on BIO tagging \citep{ratinov-roth-2009-design}, which assigns a label to each word in the sequence. This requires the input to be split into words beforehand. Word-level chunking keeps the number of negative candidates low and simplifies training, but it has two drawbacks in our setting. First, it ties supervision to word boundaries, so the model only updates the representations at those boundaries. Second, and more importantly, splitting text into words requires accurate word-segmentation models for each language, which is impractical when covering many languages and scripts.

To avoid this, we train directly on the raw input text and let the model handle segmentation itself, forming candidate spans over subword tokens rather than words. This increases the number of negative span candidates, but it removes the need for external, language-specific preprocessing. We quantify this trade-off, including its negative consequences, in \Cref{sec:pre_tokenized_text_impact_appendix} and \Cref{sec:error_analysis_appendix}.

Further, this formulation natively supports nested NER because each candidate span--label pair is scored independently. Overlapping and nested predictions therefore require no special treatment.


\begin{table*}[!ht]
\centering
\begin{tabular}{l|ccccc|c}
\toprule
& \multicolumn{5}{c|}{\textit{Backbone}} & \\
\textit{Architecture} & mDeBERTa & mmBERT & mT5 & RemBERT & XLM-R & Avg. \\
\midrule
Bi-Encoder & 0.458 & \textbf{0.489} & 0.425 & \underline{0.471} & 0.456 & 0.460 \\
Cross-Encoder & 0.494 & \textbf{0.529} & 0.427 & \underline{0.528} & 0.441 & 0.484 \\
\midrule
Avg. & 0.476 & \textbf{0.509} & 0.426 & \underline{0.499} & 0.448 & -- \\
\bottomrule
\end{tabular}
\caption{Averaged micro-F1 scores on the holdout validation benchmarks for each architecture--backbone pair. Within each row, the best and second-best scores are shown in bold and underlined, respectively.}
\label{tab:backbone_architecture_matrix}
\end{table*}

\begin{figure*}[t]
    \includegraphics[width=\linewidth]{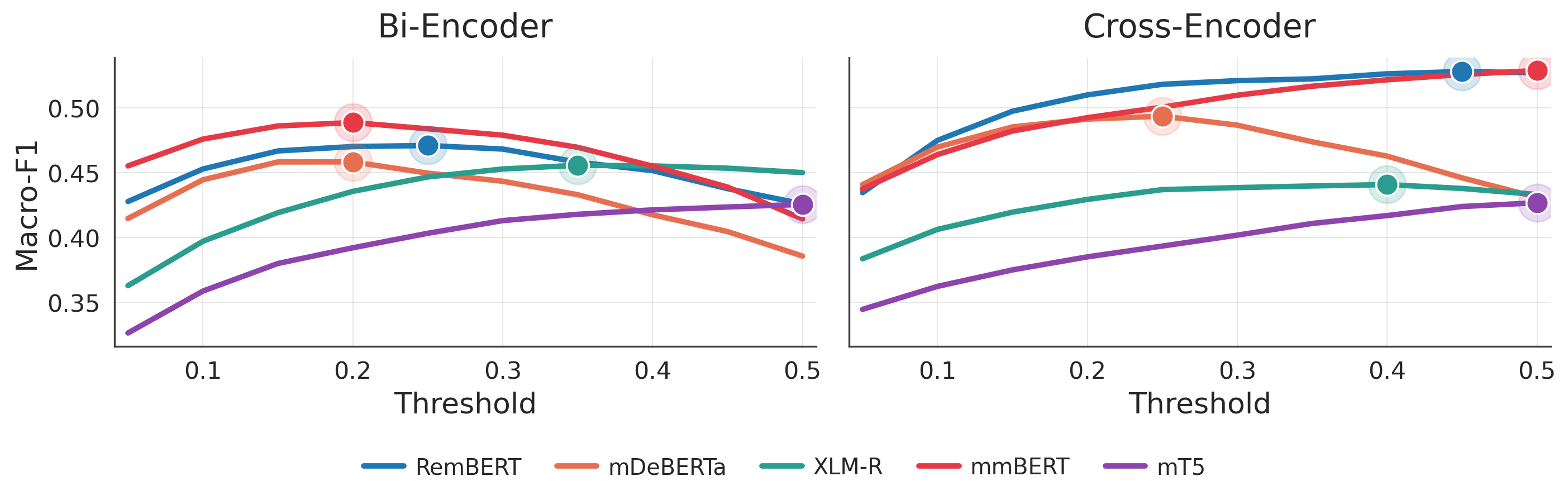}
    \caption{Average micro-F1 scores across the holdout validation benchmarks for different decision thresholds $\tau$. We observe that the optimal threshold depends mostly on the architecture (early vs.\ late fusion).}
    \label{fig:threshold_comparison}
\end{figure*}

\section{Experiments}
In the following, we evaluate each of the dimensions introduced above. We do not try every combination of the five dimensions, but instead look at one dimension at a time: we keep the best setting from the earlier steps and sweep over the next dimension.

\subsection{Experiment 1: Architecture and Backbone} \label{sec:dim1}
We start by comparing the cross-encoder and bi-encoder architectures. We train both models on the PileNER dataset (English-only, Latin script) using five multilingual transformer backbones: \textit{(i)} \texttt{xlm-roberta-base} \citep{conneau-etal-2020-unsupervised}, \textit{(ii)} \texttt{mmBERT} \citep{marone2025mmbertmodernmultilingualencoder}, \textit{(iii)} \texttt{mT5-base} \citep{xue-etal-2021-mt5}, \textit{(iv)} \texttt{mdeberta-v3-base} \citep{he2023debertav3improvingdebertausing}, and \textit{(v)} \texttt{rembert} \citep{chung2021rethinking}.

\paragraph{Hyperparameters.}~We train all models for 10k steps without early stopping, using a batch size of 12 and the AdamW optimizer \citep{loshchilov2019decoupledweightdecayregularization}. Unless stated otherwise, we use a learning rate of $3\times10^{-5}$ for all transformer backbones and MLPs, following the settings recommended in the original works. For the mT5 backbone, we use a higher learning rate of $1\times10^{-3}$, as suggested in the original paper. We fix the maximum sequence length to 512 tokens (1024 for mmBERT) and consider subword spans of up to length 30. We set the output dimension of all MLP projections to $d_{\operatorname{MLP}} = 384$ and that of the span-width embedding layer to $d_{\mathrm{width}} = 128$. For the bi-encoder setup, we use \texttt{multilingual-bert-base-uncased} \citep{devlin-etal-2019-bert} as the label encoder and represent labels using the \texttt{[CLS]} token. We use the standard binary cross-entropy loss and leave the exploration of different datasets and loss functions to later sections.

We use a holdout set of eight languages (English, German, French, Spanish, Chinese, Arabic, Hindi, and Japanese), drawn from the validation splits of MultiNERD \citep{tedeschi-navigli-2022-multinerd} and PAN-X \citep{pmlr-v119-hu20b}, with 500 samples per split. We optimize a single threshold $\tau$ on this holdout set and transfer this threshold across all languages in the later experiments.

\paragraph{Results.}~We report results on the validation set in \Cref{tab:backbone_architecture_matrix}, using the best global threshold for each architecture--backbone pair. The cross-encoder generally outperforms the bi-encoder on average (0.484 vs.\ 0.460 F1), and mmBERT is the strongest backbone overall (0.509 F1 averaged across architectures), followed closely by RemBERT (0.499 F1). For both architectures, mmBERT is the best backbone (0.489 F1 for the bi-encoder, 0.529 F1 for the cross-encoder) and RemBERT the second-best, suggesting that the relative ranking of backbones is fairly stable across architectures. Backbone choice nonetheless matters substantially: mT5 is consistently the weakest, trailing mmBERT by 0.064 F1 in the bi-encoder setting and by 0.102 F1 in the cross-encoder setting. XLM-R is the one backbone whose ranking depends on the architecture, performing reasonably in the bi-encoder setup (0.456 F1) but dropping to a low cross-encoder score (0.441 F1).

\paragraph{Threshold Selection.}~
We further evaluate model performance across a range of global decision thresholds $\tau \in \{0.05, 0.1, 0.15, \dots, 0.5\}$, in steps of $0.05$, used to convert span--label scores into final predictions. \Cref{fig:threshold_comparison} shows the averaged micro-F1 over the eight validation splits for each backbone. We highlight the optimal threshold per curve (architecture--backbone pair). We observe a consistent difference between the two architectures: the bi-encoder peaks at lower thresholds (roughly $\tau \in [0.2, 0.3]$), whereas the cross-encoder peaks at higher thresholds (roughly $\tau \in [0.4, 0.5]$). This follows from how each architecture combines span and label information, which we verify directly by scoring each candidate span--label pair on the holdout splits and comparing the resulting distributions (\Cref{sec:calibration_appendix}). The bi-encoder relies on late interaction, scoring spans and labels independently before comparing them: it assigns gold spans a mean probability of only 0.384, and just 40.5\% of them exceed 0.5, so its correct matches are best captured at lower thresholds. The cross-encoder fuses the two sources of information early, yielding sharper and better-calibrated scores (0.573 mean probability, with 62.1\% above 0.5) while pushing negatives lower, which favors higher thresholds.

\begin{figure}[t]
    \includegraphics[width=\linewidth]{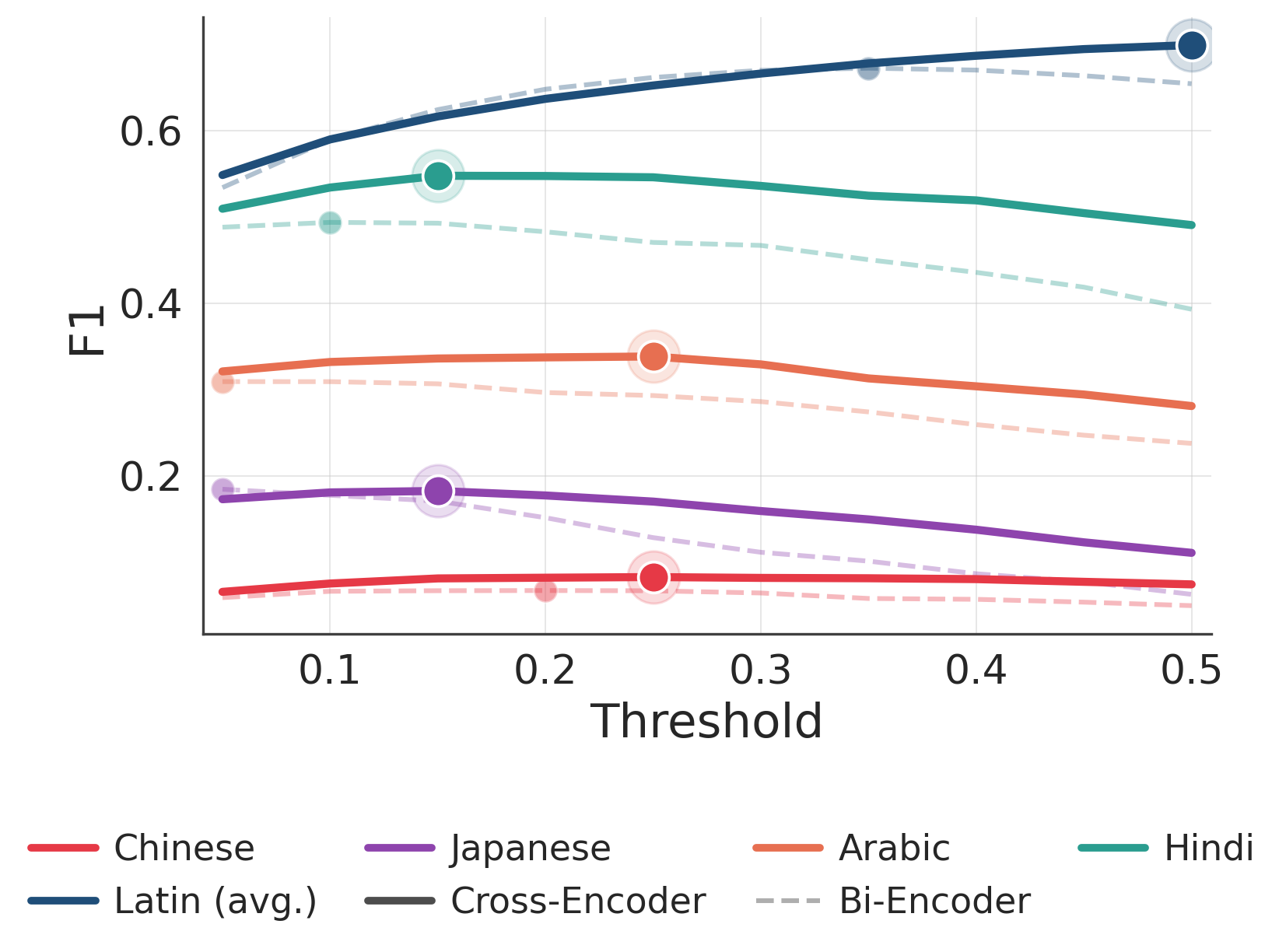}
    \caption{F1 versus decision threshold per language, macro-averaged over backbones for each architecture (cross-encoder: solid; bi-encoder: dashed). Latin (avg.) is averaged over English, German, French, and Spanish; the optimal threshold per curve is marked.}
    \label{fig:threshold_per_language_with_pilener}
\end{figure}

\begin{figure}[t]
    \includegraphics[width=0.9\linewidth]{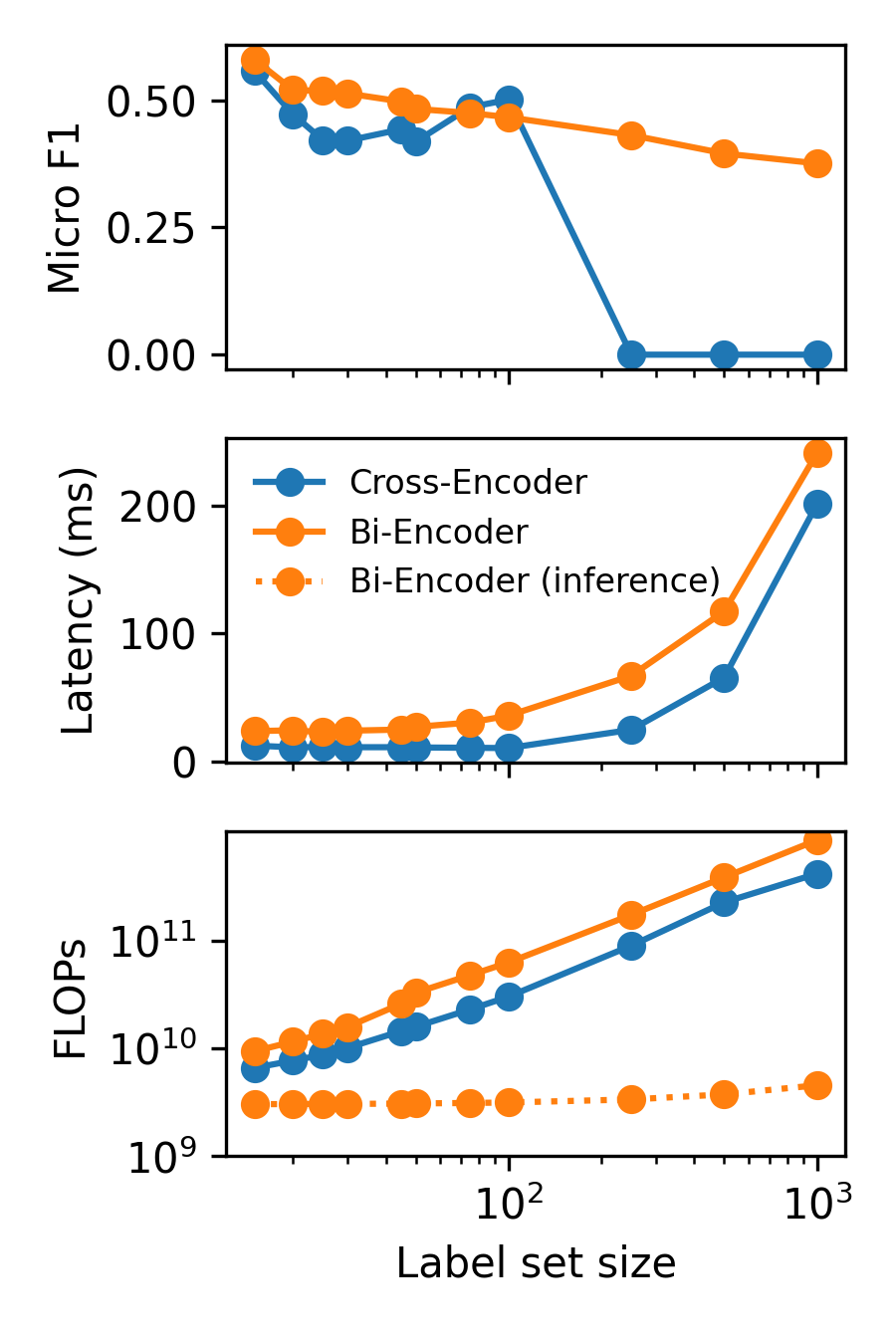}
    \caption{Micro-F1, latency, and FLOPs with increasing number of labels for cross- and bi-encoder (mmBERT). The dotted line shows bi-encoder inference FLOPs with cached label embeddings.}
    \label{fig:compute_ablation}
\end{figure}

\paragraph{Cross-encoders transfer better across scripts.}~
We further examine thresholding across different languages and scripts. We show the threshold sweep from the previous experiment grouped by script in \Cref{fig:threshold_per_language_with_pilener}. We observe that the two architectures behave almost identically on Latin-script languages (English, German, French, and Spanish), peaking at high thresholds. Non-Latin scripts, in contrast, peak at lower thresholds. We further find that subword fragmentation across scripts is what drives this observation: in \Cref{sec:fragmentation_appendix}, we show that the per-language fragmentation rates correlate with the optimal threshold, with Chinese and Japanese text producing roughly $1.1$ subwords per character and peaking at $\tau \in [0.25, 0.30]$, whereas Latin-script text produces $0.26$--$0.34$ subwords per character and peaks at $\tau \in [0.45, 0.60]$.

\begin{table*}[t]
\centering
\begin{tabular}{l|cc|cc|r}
\toprule
& \multicolumn{2}{c}{\textit{Bi-Encoder}} & \multicolumn{2}{c}{\textit{Cross-Encoder}} &  \\
\textit{Dataset} & RemBERT & mmBERT & RemBERT & mmBERT & Avg. \\
\midrule
PileNER & 0.471 & 0.489 & \underline{0.528} & \textbf{0.529} & 0.504 \\
Euro-GLiNER-x & 0.488 & \underline{0.575} & 0.534 & \textbf{0.612} & 0.552 \\
FiNERweb & 0.453 & 0.474 & \underline{0.555} & \textbf{0.575} & 0.514 \\
\bottomrule
\end{tabular}
\caption{Averaged micro-F1 on the validation splits for different training datasets. Within each row, the best and second-best scores are shown in \textbf{bold} and \underline{underlined}, respectively.}
\label{tab:lang_split_matrix}
\end{table*}

\begin{figure*}[t]
    \includegraphics[width=\linewidth]{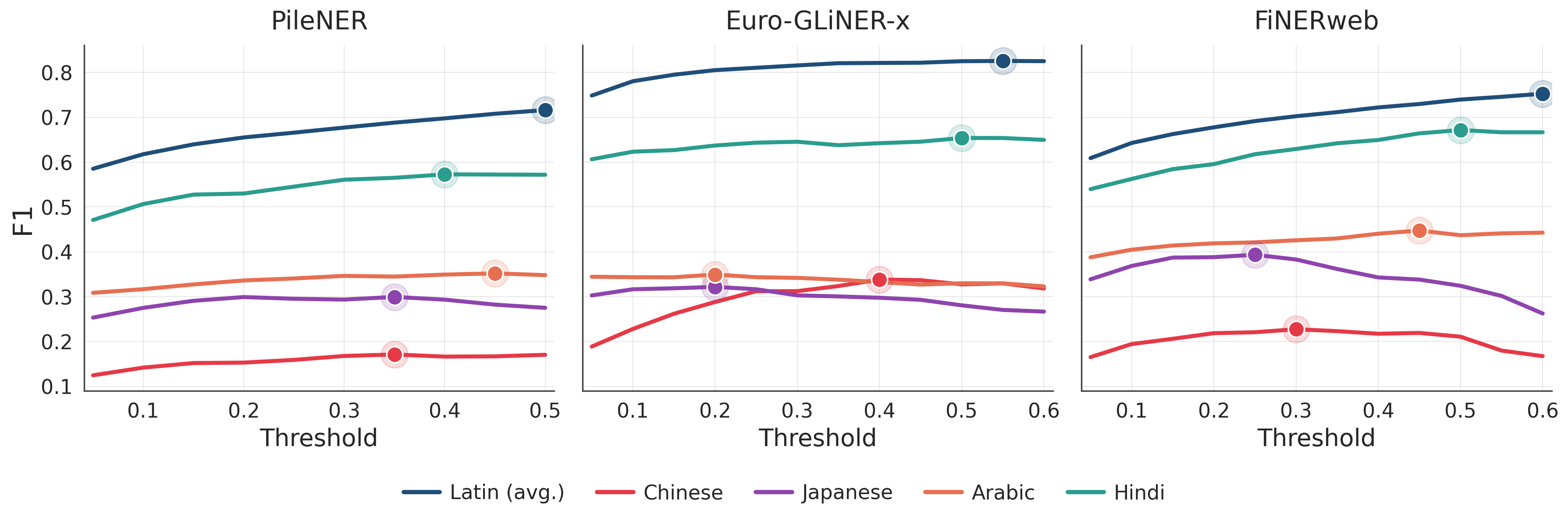}
    \caption{Macro-averaged micro-F1 scores for different decision thresholds $\tau$ and different training datasets. We observe that the optimal threshold for each language depends on the languages and scripts covered during training. We use the cross-encoder architecture with mmBERT.}
    \label{fig:threshold_comparison_multilingual_pretraining}
\end{figure*}

\paragraph{Efficiency Comparison.}~In this experiment, we compare how performance scales with the number of labels used during inference. Specifically, we use the mmBERT versions of both architectures and evaluate on the four MultiNERD validation datasets, which have 15 distinct labels. We extend this label set up to 1{,}000 label descriptions by adding the most frequent labels from PileNER, thereby introducing noise and ambiguity, such as synonyms or more fine-grained labels. In \Cref{fig:compute_ablation}, we show micro-F1, inference latency, and FLOPs as the label set grows with noisy and ambiguous labels. We find that micro-F1 decreases for both architectures as the label set grows, indicating that larger, more confusable label sets decrease performance. The cross-encoder architecture collapses beyond 250 labels: once not all labels fit in a single forward pass, logits across batches become incomparable. The bi-encoder naturally overcomes this issue, as its scores are computed independently of the batch composition. In terms of cost, the bi-encoder has higher training FLOPs due to its two encoders, but at inference its label embeddings can be cached and reused across inputs, making inference roughly linear in the label set size and far cheaper. Bi-encoder latency is nearly identical with and without caching, so we omit the latter curve for readability.

\subsection{Experiment 2: Multilingual Fine-Tuning}

We now analyze the effect of the fine-tuning dataset. We consider three datasets: \textit{(i)} PileNER \citep{zhou2024universalnertargeteddistillationlarge}, which is English-only; \textit{(ii)} Euro-GLiNER-x \footnote{\url{https://huggingface.co/datasets/knowledgator/gliner-multilingual-synthetic}}, which covers 12 Indo-European languages in Latin script; and \textit{(iii)} FiNERweb \citep{golde2025finerwebdatasetsartifactsscalable}, which spans 91 languages and 25 scripts. All three follow a long-tail distribution of entity types. These experiments aim to answer to what extent fine-tuning a multilingual backbone on English-only data is sufficient, and whether multilingual training data is required.

\paragraph{Interpreting the comparison.}~The three corpora differ in language coverage, but also in their underlying generation procedure. Isolating coverage as a single variable is therefore not strictly possible. However, all runs use the same architecture, backbone, objective, optimizer settings, and step budget, so the comparison is at least held fixed on the training side. We further conduct a basic lexical and semantic comparison between the training and validation data in \Cref{sec:similarity_appendix} to better interpret the findings of this section. Corpus size and annotation quality likely contribute to the absolute numbers, and we therefore read these results as evidence about corpora as they exist rather than about coverage in isolation. One factor we can quantify directly is how many subword embeddings each corpus actually updates during training. We tokenize each dataset with the XLM-R tokenizer and compute the fraction of subword embeddings that receive gradient updates (\Cref{sec:subword_updates}). FiNERweb updates over 94\% of all subword embeddings, compared to only 38.55\% for English-only PileNER, indicating that much of the multilingual vocabulary is never trained under English-only fine-tuning.

\paragraph{Results.}~We report averaged micro-F1 scores per training dataset in \Cref{tab:lang_split_matrix}. The cross-encoder architecture outperforms the bi-encoder across all datasets and backbones, and mmBERT is again the strongest backbone. We find that multilingual training helps overall: under a matched budget, the best result for each architecture is obtained with a multilingual dataset rather than with PileNER. Euro-GLiNER-x is the strongest dataset on average (0.552), improving over PileNER for both architectures (cross-encoder with mmBERT: 0.529 $\rightarrow$ 0.612). FiNERweb also improves the cross-encoder over PileNER (cross-encoder with mmBERT: 0.529 $\rightarrow$ 0.575) but, unlike Euro-GLiNER-x, does not help the bi-encoder, where it falls slightly below English-only training.

\paragraph{Per-language analysis.}~\Cref{fig:threshold_comparison_multilingual_pretraining} breaks down F1 by language for the cross-encoder with mmBERT. The gains from multilingual training depend on the script. On Latin-script languages, FiNERweb does not beat the other datasets and even falls slightly behind, likely because its broad coverage leaves less room for Latin-script data. On non-Latin scripts, it is clearly ahead: Arabic, Hindi, and Japanese all score higher than under English- or Latin-only training. This split is what we would expect if coverage is the deciding factor, since corpus size and annotation quality should not push Latin and non-Latin scripts in opposite directions. Broad coverage thus costs a little on Latin scripts and gains a lot on the rest.

\subsection{Experiment 3: Loss Functions}~In this experiment, we compare different loss functions: \textit{(i)} binary cross-entropy (BCE) (with positive upweighting), \textit{(ii)} focal loss with varying $\alpha$ and $\gamma$, \textit{(iii)} contrastive loss and \textit{(iv)} dice loss. We provide details on the loss formulations in \Cref{sec:loss_functions_appendix} and use the cross-encoder with mmBERT trained on FiNERweb.

\begin{table}[t]
\centering
\begin{tabular}{lc}
\toprule
\textit{Loss Function} & \textit{F1 Score} \\
\midrule
Baseline (BCE) & 0.575 \\
\midrule
\textit{BCE + Pos. Weight $\lambda$} \\
\quad $\lambda = 10.0$ & 0.530 \\
\quad $\lambda = 100.0$ & 0.432 \\
\midrule
\textit{Focal Loss} \\
\quad $\alpha=0.75, \gamma=1.0$ & \textbf{0.585} \\
\quad $\alpha=0.25, \gamma=2.0$ & 0.573 \\
\midrule
\textit{Contrastive Loss} \\
\quad $\alpha=0.5, \beta=0.5$ & 0.046 \\
\midrule
\textit{Dice Loss} & \underline{0.577} \\
\bottomrule
\end{tabular}
\caption{Averaged micro-F1 on the validation splits for different loss functions (cross-encoder, mmBERT, FiNERweb).}
\label{tab:loss_function_ce_mmbert}
\end{table}

\paragraph{Results.}~We report results in \Cref{tab:loss_function_ce_mmbert}. First, we observe that upweighting positives does not help, but instead degrades performance. Dice loss roughly matches BCE (0.577 vs.\ 0.575), and contrastive loss fails to train competitively in our setting (0.046). Focal loss gives the best holdout result (0.585 with $\alpha{=}0.75, \gamma{=}1.0$), a marginal improvement over BCE. One hypothesis for this gain is the breadth of label descriptions in FiNERweb: with many overlapping and fine-grained labels, down-weighting easy negatives may help.

\begin{table*}[t]
\centering
\begin{tabular}{l|ccccccc|r}
\toprule
& \multicolumn{7}{c}{\textit{Datasets}} & \\ 
& \small{Dynamic-} & \small{Masakha-} & \multicolumn{2}{c}{\small{MultiCoNER}} & \small{Multi-} & \small{PAN-X} & \small{UNER} & \small{Avg.} \\
\textit{Model} & \small{NER} & \small{NER} & \small{v1} & \small{v2} & \small{NERD} & & \\
\midrule
\textit{LLMs} \\
GPT-5 & 0.204 & 0.388 & 0.267 & 0.133 & 0.496 & 0.477 & 0.468 & 0.347 \\ 
Qwen3-32B & 0.365 & 0.535 & 0.419 & 0.349 & 0.617 & 0.554 & 0.681 & 0.503 \\ 
Gemma3-27B & 0.434 & 0.594 & 0.428 & 0.373 & 0.646 & 0.584 & 0.742 & 0.543 \\ 
\midrule
\textit{Universal NER Models} \\
WikiNeural & 0.001 & 0.307 & 0.097 & 0.013 & 0.652 & 0.403 & 0.506 & 0.283 \\ 
GLiNER-multi-v2.1 & 0.291 & 0.480 & 0.366 & 0.238 & 0.533 & 0.532 & 0.551 & 0.427 \\ 
GLiNER-x-base & 0.187 & 0.559 & 0.329 & 0.210 & 0.582 & 0.509 & 0.644 & 0.431 \\ 
\midrule
\methodname{} \small \textsc{(Bi-Enc.)} \\
\quad w/ RemBERT & 0.166 & 0.541 & 0.351 & 0.156 & 0.595 & 0.508 & 0.704 & 0.432 \\
\quad w/ mmBERT & 0.251 & 0.485 & 0.355 & 0.182 & 0.576 & 0.479 & 0.661 & 0.427 \\
\midrule
\methodname{} \small \textsc{(Cross-Enc.)} \\
\quad w/ RemBERT & 0.354 & 0.573 & 0.347 & 0.294 & 0.636 & 0.436 & 0.540 & 0.454 \\
\quad w/ mmBERT & 0.300 & 0.542 & 0.414 & 0.249 & 0.653 & 0.602 & 0.660 & 0.489 \\
\quad w/ mmBERT, focal & 0.321 & 0.537 & 0.401 & 0.240 & 0.649 & 0.596 & 0.680 & 0.489 \\
\bottomrule
\end{tabular}
\caption{Macro-averaged F1 scores across NER benchmarks. All \methodname{} rows use a single global threshold; unless marked otherwise they are trained with BCE, and the focal-loss row is matched to it at 30k training steps.}
\label{tab:baseline_models_small}
\end{table*}

\section{Combining The Insights}
\label{sec:combining}
We finally combine our findings to obtain \methodname{}, training our best architecture--backbone combination on FiNERweb for 30k steps without early stopping, using the mmBERT backbone and binary cross-entropy loss. We choose FiNERweb for its broad language and script coverage, and BCE because the more advanced loss functions yielded only marginal gains. We retain BCE rather than the nominally better focal loss because the two tie across the full benchmark suite (\Cref{tab:baseline_models_small}), so ``optimized'' here denotes the best performance--simplicity trade-off rather than the literal union of every top-scoring choice. Based on our threshold analysis, we adopt a single global threshold per architecture, $\tau=0.3$ for the cross-encoder and $\tau=0.2$ for the bi-encoder, choosing values at the lower end of the optimal range since non-Latin scripts peak at lower thresholds and we evaluate across many languages and splits. To situate \methodname{} with respect to related work, we compare it against two families of baselines: general-purpose multilingual LLMs and dedicated universal NER models. The former includes GPT-5\footnote{\url{https://platform.openai.com/docs/models/gpt-5}}, Qwen3-32B \citep{yang2025qwen3technicalreport}, and Gemma3-27B \citep{gemmateam2025gemma3technicalreport}; the latter includes WikiNeural \citep{tedeschi-etal-2021-wikineural-combined} and two multilingual GLiNER variants \citep{zaratiana-etal-2024-gliner}.

\paragraph{Evaluation Benchmarks.}~
We use seven multilingual datasets for evaluation: DynamicNER \citep{luo-etal-2025-dynamicner}, UNER \citep{mayhew-etal-2024-universal}, MasakhaNER 2.0 \citep{adelani-etal-2022-masakhaner}, MultiNERD \citep{tedeschi-navigli-2022-multinerd}, MultiCoNER v1 \citep{malmasi-etal-2022-multiconer} and v2 \citep{fetahu-etal-2023-multiconer} and PAN-X \citep{pmlr-v119-hu20b}. We provide an overview of the number of languages and label set sizes in \Cref{tab:ner-datasets}. As our evaluation covers 250 test splits in total, we limit each test split to 1{,}000 examples when it is larger. We report all results using micro-averaged F1 across languages within each dataset, and macro-averaged F1 across datasets when reporting aggregate results.

\paragraph{Annotation conventions and class imbalance.}~These benchmarks were annotated independently and differ in label definitions and span-boundary conventions, and their positive-class rates vary by more than an order of magnitude across languages (\Cref{fig:positive_negative_ratios_tokenize}). This bounds what any model can score and applies equally to all systems we compare; \Cref{sec:error_analysis_appendix} gives a concrete case in which our predictions are penalized by a boundary convention rather than being modeling errors. A systematic study of annotation noise across these benchmarks is a valuable but separate contribution.

\paragraph{Results.}~We report all baselines and \methodname{} in \Cref{tab:baseline_models_small}. All results for \methodname{} use the single global threshold described above, i.e.\ the deployable setting that requires no target-language validation data. Our best cross-encoder reaches 0.489 F1, the strongest result among similarly sized task-specific models, beating GLiNER-x-base (0.431) by 5.7\,pp. It remains 1.4\,pp behind Qwen3-32B (0.503) and 5.4\,pp behind Gemma3-27B (0.543), despite being roughly $90\times$ smaller. Training the identical configuration with focal loss instead of BCE gives the same average of 0.489, confirming that the objective is not what drives this result.

The cross-encoder outperforms the bi-encoder at scale (0.489 vs.\ 0.427 for mmBERT), unlike our earlier holdout experiments where the two were comparable. This is consistent with its early fusion of span and label information capturing language-specific patterns, such as the varying positive-to-negative span ratio across languages: stratifying the per-language F1 gap by the negative-to-positive candidate ratio shows that the cross-encoder advantage is largest where positive spans are dense and smallest where they are sparse (\Cref{sec:negpos_appendix}).

Among LLMs, Gemma3-27B is strongest overall (0.543), ahead of Qwen3-32B (0.503), while GPT-5 is surprisingly weak (0.347). We select Gemma3 deliberately, as it is one of the teachers used to annotate FiNERweb, which makes the comparison a measure of how much teacher performance a $90\times$ smaller student recovers. Absolute F1, however, is only one side of the picture. At 18\,ms per example and $1.4$\,GB of VRAM, \methodname{} runs 50--80$\times$ faster than a 4B-parameter baseline that it outperforms by 15--31\,pp, and roughly two orders of magnitude faster than the 27--32B models leading the table (\Cref{sec:efficiency_appendix}). Our claim is therefore about the trade-off rather than the maximum: at the scale at which universal NER models are deployed, a task-specific model comes within a few points of models that are far more expensive to run.

\section{Related Work}

\paragraph{Natural Language Prompting with LLMs.}~The advent of increasingly capable autoregressive language models has introduced a new paradigm based on natural language prompting \citep{brown2020languagemodelsfewshotlearners,schick-schutze-2021-just,min-etal-2022-rethinking}, effectively replacing the fixed output layer. This paradigm has been widely applied to information extraction tasks, including text classification \citep{halder-etal-2020-task,sun-etal-2023-text}, entity linking \citep{decao2021autoregressiveentityretrieval,ding-etal-2024-chatel}, and named entity recognition \citep{huang-etal-2022-copner,ashok2023promptnerpromptingnamedentity}, as well as joint modeling across multiple tasks \citep{wang2023instructuiemultitaskinstructiontuning}.

\paragraph{Knowledge Distillation from Synthetic Datasets.}~A major drawback of large language models is their substantial computational cost. Consequently, more recent work no longer relies on the autoregressive generation process at inference time, but instead adopts a knowledge distillation framework \citep{hinton2015distillingknowledgeneuralnetwork}. In this setup, LLMs serve as teachers to produce annotated datasets in a one-off process \citep{ye-etal-2022-zerogen}. This approach has been successfully applied to named entity recognition \citep{zhou2024universalnertargeteddistillationlarge,bogdanov-etal-2024-nuner,golde2025finerwebdatasetsartifactsscalable}.

\paragraph{Universal NER.}~Recent work such as Binder \citep{zhang2023optimizing} and NuNER \citep{bogdanov-etal-2024-nuner} employs bi-encoder architectures, whereas USM \citep{lou2023universalinformationextractionunified} and GLiNER \citep{zaratiana-etal-2024-gliner} rely on cross-encoders. However, these models differ substantially in their loss functions, transformer backbones, and training data. In contrast, our work compares these design choices in a controlled experimental setting. Finally, our work follows the same research direction as \citet{huang-etal-2019-matters}, who use pre-transformer architectures, train on at most 14 languages, and evaluate on 4. We train on up to 91 languages and evaluate on 250+ test splits with current universal-NER architectures, and find a picture more specific than ``multilingual fine-tuning helps'': it helps unevenly across scripts (\Cref{fig:threshold_comparison_multilingual_pretraining}), does not help the bi-encoder at all with FiNERweb (\Cref{tab:lang_split_matrix}), and backbone rankings can flip with the architecture (\Cref{tab:backbone_architecture_matrix}) --- none of which follows from first principles.

\section{Conclusion}~In this work, we systematically explored the design space of prior approaches to universal NER under a staged, controlled protocol. Based on these insights, we derived \methodname{}, a new state-of-the-art universal NER model that generalizes to more than 100 languages. Our results indicate that \textit{(i)} multilingual training data is essential for universal NER, \textit{(ii)} the effectiveness of a transformer backbone strongly depends on the chosen architecture, \textit{(iii)} cross-encoders outperform bi-encoders at scale in terms of generalization, \textit{(iv)} a simple binary cross-entropy loss is sufficient, and \textit{(v)} threshold selection plays a critical role in universal, multilingual NER. Under a single global threshold, the resulting model stays within a few points of generative models two orders of magnitude larger while running one to two orders of magnitude faster, which we regard as the practically relevant operating point.

\clearpage

\section*{Limitations}
\noindent\textbf{Pretrained model and data constraints.}
Our proposed approach is limited by the availability and quality of pretrained multilingual language models and training data, as the design of \methodname{} is empirically derived from existing architectures and datasets. In this work, the largest number of languages covered by a single training dataset is 91.

\noindent\textbf{Staged rather than exhaustive search.}
Fully crossing all five design dimensions would require several hundred training runs, so we proceed in stages: each dimension is varied while the choices preferred in the earlier stages are held fixed. Concretely, the loss-function comparison is run only with the cross-encoder, the mmBERT backbone, and FiNERweb, and the threshold analysis is conditioned on the corresponding checkpoints. Our conclusions are therefore statements about the configurations we tested, and a different objective could in principle interact differently with another backbone or corpus. Two observations mitigate this. First, where we do cross dimensions, the conclusions are stable: the cross-encoder is ahead of the bi-encoder for every backbone--dataset pair in \Cref{tab:lang_split_matrix}, and mmBERT is the best backbone under both architectures in \Cref{tab:backbone_architecture_matrix}. Second, the one interaction that we do observe, the architecture-dependent ranking of XLM-R, is reported explicitly rather than averaged away. Still, a full factorial study remains a valuable direction for future work.

\noindent\textbf{Confounds in the dataset comparison.}
The three fine-tuning corpora differ in language coverage but also in size, annotation pipeline, and label granularity. We hold the architecture, backbone, objective, and step budget fixed across them and rule out distributional proximity to the evaluation data as an explanation (\Cref{sec:similarity_appendix}), but we cannot attribute the observed differences to multilingual coverage alone. An experiment that varies language coverage while holding the annotation pipeline fixed would be needed to establish that claim conclusively.

\noindent\textbf{Annotation noise and cross-dataset scheme mismatch.}
NER benchmarks are known to differ in span-boundary conventions and to contain annotation noise, which affects cross-dataset zero-shot evaluation for all models we compare. \Cref{sec:error_analysis_appendix} documents one concrete instance of this on Shona, where linguistically plausible predictions are penalized by the annotation convention rather than being modeling errors. We do not attempt to quantify how much of the remaining gap across the 250 test splits is attributable to such mismatches; doing so would require a systematic re-annotation study.

\noindent\textbf{Language coverage.}
Although we evaluate \methodname{} on more than 150 languages, only a subset of these languages is represented in the training data. As a result, performance may degrade for languages outside the investigated training scope, in particular for low-resource languages or scripts that are underrepresented in the pretrained models and/or the training data.

\noindent\textbf{Label semantics.}
The evaluation benchmarks considered in this work use label sets with distinct class boundaries such as ``person'' and ``location''. We do not explicitly investigate the effect of semantic similarity between entity labels, e.g.\ ``person'' and ``human''. Since we train the model to treat such labels as distinct concepts, this may contribute to the observed performance differences when training with labels in the target language.

\noindent\textbf{Threshold Selection.}
We find that threshold selection for zero-shot NER is highly language-dependent, and therefore select the threshold on a holdout validation set. As a consequence, the performance gap to in-domain fine-tuned models can remain substantial, since label definitions and, in particular, span-boundary conventions differ across datasets. As a result, our model can serve as a suitable starting point for further fine-tuning on target datasets, where dataset-specific label and boundary definitions can be incorporated.

\section*{Acknowledgments}
We thank all reviewers for their valuable comments. Jonas Golde is supported by the Bundesministerium für Bildung und Forschung (BMBF) as part of the project ``FewTuRe'' (project number 01IS24020). Alan Akbik and Patrick Haller are supported by the Deutsche Forschungsgemeinschaft (DFG, German Research Foundation) under Emmy Noether grant ``Eidetic Representations of Natural Language'' (project number 448414230). Further, Alan Akbik is supported under Germany’s Excellence Strategy ``Science of Intelligence'' (EXC 2002/1, project number 390523135).


\bibliography{custom}

@inproceedings{lample-etal-2016-neural,
    title = "Neural Architectures for Named Entity Recognition",
    author = "Lample, Guillaume  and
      Ballesteros, Miguel  and
      Subramanian, Sandeep  and
      Kawakami, Kazuya  and
      Dyer, Chris",
    editor = "Knight, Kevin  and
      Nenkova, Ani  and
      Rambow, Owen",
    booktitle = "Proceedings of the 2016 Conference of the North {A}merican Chapter of the Association for Computational Linguistics: Human Language Technologies",
    month = jun,
    year = "2016",
    address = "San Diego, California",
    publisher = "Association for Computational Linguistics",
    url = "https://aclanthology.org/N16-1030/",
    doi = "10.18653/v1/N16-1030",
    pages = "260--270"
}

@inproceedings{li-etal-2020-dice,
    title = "Dice Loss for Data-imbalanced {NLP} Tasks",
    author = "Li, Xiaoya  and
      Sun, Xiaofei  and
      Meng, Yuxian  and
      Liang, Junjun  and
      Wu, Fei  and
      Li, Jiwei",
    editor = "Jurafsky, Dan  and
      Chai, Joyce  and
      Schluter, Natalie  and
      Tetreault, Joel",
    booktitle = "Proceedings of the 58th Annual Meeting of the Association for Computational Linguistics",
    month = jul,
    year = "2020",
    address = "Online",
    publisher = "Association for Computational Linguistics",
    url = "https://aclanthology.org/2020.acl-main.45/",
    doi = "10.18653/v1/2020.acl-main.45",
    pages = "465--476"
}

@inproceedings{akbik-etal-2018-contextual,
    title = "Contextual String Embeddings for Sequence Labeling",
    author = "Akbik, Alan  and
      Blythe, Duncan  and
      Vollgraf, Roland",
    editor = "Bender, Emily M.  and
      Derczynski, Leon  and
      Isabelle, Pierre",
    booktitle = "Proceedings of the 27th International Conference on Computational Linguistics",
    month = aug,
    year = "2018",
    address = "Santa Fe, New Mexico, USA",
    publisher = "Association for Computational Linguistics",
    url = "https://aclanthology.org/C18-1139/",
    pages = "1638--1649"
}

@article{zhong2023comprehensive,
  title={A comprehensive survey on automatic knowledge graph construction},
  author={Zhong, Lingfeng and Wu, Jia and Li, Qian and Peng, Hao and Wu, Xindong},
  journal={ACM Computing Surveys},
  volume={56},
  number={4},
  pages={1--62},
  year={2023},
  publisher={ACM New York, NY}
}

@inproceedings{perot-etal-2024-lmdx,
    title = "{LMDX}: Language Model-based Document Information Extraction and Localization",
    author = "Perot, Vincent  and
      Kang, Kai  and
      Luisier, Florian  and
      Su, Guolong  and
      Sun, Xiaoyu  and
      Boppana, Ramya Sree  and
      Wang, Zilong  and
      Wang, Zifeng  and
      Mu, Jiaqi  and
      Zhang, Hao  and
      Lee, Chen-Yu  and
      Hua, Nan",
    editor = "Ku, Lun-Wei  and
      Martins, Andre  and
      Srikumar, Vivek",
    booktitle = "Findings of the Association for Computational Linguistics: ACL 2024",
    month = aug,
    year = "2024",
    address = "Bangkok, Thailand",
    publisher = "Association for Computational Linguistics",
    url = "https://aclanthology.org/2024.findings-acl.899/",
    doi = "10.18653/v1/2024.findings-acl.899",
    pages = "15140--15168"
}

@inproceedings{arsenyan-etal-2024-large,
    title = "Large Language Models for Biomedical Knowledge Graph Construction: Information extraction from {EMR} notes",
    author = "Arsenyan, Vahan  and
      Bughdaryan, Spartak  and
      Shaya, Fadi  and
      Small, Kent Wilson  and
      Shahnazaryan, Davit",
    editor = "Demner-Fushman, Dina  and
      Ananiadou, Sophia  and
      Miwa, Makoto  and
      Roberts, Kirk  and
      Tsujii, Junichi",
    booktitle = "Proceedings of the 23rd Workshop on Biomedical Natural Language Processing",
    month = aug,
    year = "2024",
    address = "Bangkok, Thailand",
    publisher = "Association for Computational Linguistics",
    url = "https://aclanthology.org/2024.bionlp-1.23/",
    doi = "10.18653/v1/2024.bionlp-1.23",
    pages = "295--317"
}

@inproceedings{shachar-etal-2025-ner,
    title = "{NER} Retriever: Zero-Shot Named Entity Retrieval with Type-Aware Embeddings",
    author = "Shachar, Or  and
      Katz, Uri  and
      Goldberg, Yoav  and
      Glickman, Oren",
    editor = "Christodoulopoulos, Christos  and
      Chakraborty, Tanmoy  and
      Rose, Carolyn  and
      Peng, Violet",
    booktitle = "Findings of the Association for Computational Linguistics: EMNLP 2025",
    month = nov,
    year = "2025",
    address = "Suzhou, China",
    publisher = "Association for Computational Linguistics",
    url = "https://aclanthology.org/2025.findings-emnlp.597/",
    doi = "10.18653/v1/2025.findings-emnlp.597",
    pages = "11175--11186",
    ISBN = "979-8-89176-335-7"
}

@misc{grattafiori2024llama3herdmodels,
      title={The Llama 3 Herd of Models}, 
      author={Aaron Grattafiori and Abhimanyu Dubey and Abhinav Jauhri and Abhinav Pandey and Abhishek Kadian and Ahmad Al-Dahle and Aiesha Letman and Akhil Mathur and Alan Schelten and Alex Vaughan and Amy Yang and Angela Fan and Anirudh Goyal and Anthony Hartshorn and Aobo Yang and Archi Mitra and Archie Sravankumar and Artem Korenev and Arthur Hinsvark and Arun Rao and Aston Zhang and Aurelien Rodriguez and Austen Gregerson and Ava Spataru and Baptiste Roziere and Bethany Biron and Binh Tang and Bobbie Chern and Charlotte Caucheteux and Chaya Nayak and Chloe Bi and Chris Marra and Chris McConnell and Christian Keller and Christophe Touret and Chunyang Wu and Corinne Wong and Cristian Canton Ferrer and Cyrus Nikolaidis and Damien Allonsius and Daniel Song and Danielle Pintz and Danny Livshits and Danny Wyatt and David Esiobu and Dhruv Choudhary and Dhruv Mahajan and Diego Garcia-Olano and Diego Perino and Dieuwke Hupkes and Egor Lakomkin and Ehab AlBadawy and Elina Lobanova and Emily Dinan and Eric Michael Smith and Filip Radenovic and Francisco Guzmán and Frank Zhang and Gabriel Synnaeve and Gabrielle Lee and Georgia Lewis Anderson and Govind Thattai and Graeme Nail and Gregoire Mialon and Guan Pang and Guillem Cucurell and Hailey Nguyen and Hannah Korevaar and Hu Xu and Hugo Touvron and Iliyan Zarov and Imanol Arrieta Ibarra and Isabel Kloumann and Ishan Misra and Ivan Evtimov and Jack Zhang and Jade Copet and Jaewon Lee and Jan Geffert and Jana Vranes and Jason Park and Jay Mahadeokar and Jeet Shah and Jelmer van der Linde and Jennifer Billock and Jenny Hong and Jenya Lee and Jeremy Fu and Jianfeng Chi and Jianyu Huang and Jiawen Liu and Jie Wang and Jiecao Yu and Joanna Bitton and Joe Spisak and Jongsoo Park and Joseph Rocca and Joshua Johnstun and Joshua Saxe and Junteng Jia and Kalyan Vasuden Alwala and Karthik Prasad and Kartikeya Upasani and Kate Plawiak and Ke Li and Kenneth Heafield and Kevin Stone and Khalid El-Arini and Krithika Iyer and Kshitiz Malik and Kuenley Chiu and Kunal Bhalla and Kushal Lakhotia and Lauren Rantala-Yeary and Laurens van der Maaten and Lawrence Chen and Liang Tan and Liz Jenkins and Louis Martin and Lovish Madaan and Lubo Malo and Lukas Blecher and Lukas Landzaat and Luke de Oliveira and Madeline Muzzi and Mahesh Pasupuleti and Mannat Singh and Manohar Paluri and Marcin Kardas and Maria Tsimpoukelli and Mathew Oldham and Mathieu Rita and Maya Pavlova and Melanie Kambadur and Mike Lewis and Min Si and Mitesh Kumar Singh and Mona Hassan and Naman Goyal and Narjes Torabi and Nikolay Bashlykov and Nikolay Bogoychev and Niladri Chatterji and Ning Zhang and Olivier Duchenne and Onur Çelebi and Patrick Alrassy and Pengchuan Zhang and Pengwei Li and Petar Vasic and Peter Weng and Prajjwal Bhargava and Pratik Dubal and Praveen Krishnan and Punit Singh Koura and Puxin Xu and Qing He and Qingxiao Dong and Ragavan Srinivasan and Raj Ganapathy and Ramon Calderer and Ricardo Silveira Cabral and Robert Stojnic and Roberta Raileanu and Rohan Maheswari and Rohit Girdhar and Rohit Patel and Romain Sauvestre and Ronnie Polidoro and Roshan Sumbaly and Ross Taylor and Ruan Silva and Rui Hou and Rui Wang and Saghar Hosseini and Sahana Chennabasappa and Sanjay Singh and Sean Bell and Seohyun Sonia Kim and Sergey Edunov and Shaoliang Nie and Sharan Narang and Sharath Raparthy and Sheng Shen and Shengye Wan and Shruti Bhosale and Shun Zhang and Simon Vandenhende and Soumya Batra and Spencer Whitman and Sten Sootla and Stephane Collot and Suchin Gururangan and Sydney Borodinsky and Tamar Herman and Tara Fowler and Tarek Sheasha and Thomas Georgiou and Thomas Scialom and Tobias Speckbacher and Todor Mihaylov and Tong Xiao and Ujjwal Karn and Vedanuj Goswami and Vibhor Gupta and Vignesh Ramanathan and Viktor Kerkez and Vincent Gonguet and Virginie Do and Vish Vogeti and Vítor Albiero and Vladan Petrovic and Weiwei Chu and Wenhan Xiong and Wenyin Fu and Whitney Meers and Xavier Martinet and Xiaodong Wang and Xiaofang Wang and Xiaoqing Ellen Tan and Xide Xia and Xinfeng Xie and Xuchao Jia and Xuewei Wang and Yaelle Goldschlag and Yashesh Gaur and Yasmine Babaei and Yi Wen and Yiwen Song and Yuchen Zhang and Yue Li and Yuning Mao and Zacharie Delpierre Coudert and Zheng Yan and Zhengxing Chen and Zoe Papakipos and Aaditya Singh and Aayushi Srivastava and Abha Jain and Adam Kelsey and Adam Shajnfeld and Adithya Gangidi and Adolfo Victoria and Ahuva Goldstand and Ajay Menon and Ajay Sharma and Alex Boesenberg and Alexei Baevski and Allie Feinstein and Amanda Kallet and Amit Sangani and Amos Teo and Anam Yunus and Andrei Lupu and Andres Alvarado and Andrew Caples and Andrew Gu and Andrew Ho and Andrew Poulton and Andrew Ryan and Ankit Ramchandani and Annie Dong and Annie Franco and Anuj Goyal and Aparajita Saraf and Arkabandhu Chowdhury and Ashley Gabriel and Ashwin Bharambe and Assaf Eisenman and Azadeh Yazdan and Beau James and Ben Maurer and Benjamin Leonhardi and Bernie Huang and Beth Loyd and Beto De Paola and Bhargavi Paranjape and Bing Liu and Bo Wu and Boyu Ni and Braden Hancock and Bram Wasti and Brandon Spence and Brani Stojkovic and Brian Gamido and Britt Montalvo and Carl Parker and Carly Burton and Catalina Mejia and Ce Liu and Changhan Wang and Changkyu Kim and Chao Zhou and Chester Hu and Ching-Hsiang Chu and Chris Cai and Chris Tindal and Christoph Feichtenhofer and Cynthia Gao and Damon Civin and Dana Beaty and Daniel Kreymer and Daniel Li and David Adkins and David Xu and Davide Testuggine and Delia David and Devi Parikh and Diana Liskovich and Didem Foss and Dingkang Wang and Duc Le and Dustin Holland and Edward Dowling and Eissa Jamil and Elaine Montgomery and Eleonora Presani and Emily Hahn and Emily Wood and Eric-Tuan Le and Erik Brinkman and Esteban Arcaute and Evan Dunbar and Evan Smothers and Fei Sun and Felix Kreuk and Feng Tian and Filippos Kokkinos and Firat Ozgenel and Francesco Caggioni and Frank Kanayet and Frank Seide and Gabriela Medina Florez and Gabriella Schwarz and Gada Badeer and Georgia Swee and Gil Halpern and Grant Herman and Grigory Sizov and Guangyi and Zhang and Guna Lakshminarayanan and Hakan Inan and Hamid Shojanazeri and Han Zou and Hannah Wang and Hanwen Zha and Haroun Habeeb and Harrison Rudolph and Helen Suk and Henry Aspegren and Hunter Goldman and Hongyuan Zhan and Ibrahim Damlaj and Igor Molybog and Igor Tufanov and Ilias Leontiadis and Irina-Elena Veliche and Itai Gat and Jake Weissman and James Geboski and James Kohli and Janice Lam and Japhet Asher and Jean-Baptiste Gaya and Jeff Marcus and Jeff Tang and Jennifer Chan and Jenny Zhen and Jeremy Reizenstein and Jeremy Teboul and Jessica Zhong and Jian Jin and Jingyi Yang and Joe Cummings and Jon Carvill and Jon Shepard and Jonathan McPhie and Jonathan Torres and Josh Ginsburg and Junjie Wang and Kai Wu and Kam Hou U and Karan Saxena and Kartikay Khandelwal and Katayoun Zand and Kathy Matosich and Kaushik Veeraraghavan and Kelly Michelena and Keqian Li and Kiran Jagadeesh and Kun Huang and Kunal Chawla and Kyle Huang and Lailin Chen and Lakshya Garg and Lavender A and Leandro Silva and Lee Bell and Lei Zhang and Liangpeng Guo and Licheng Yu and Liron Moshkovich and Luca Wehrstedt and Madian Khabsa and Manav Avalani and Manish Bhatt and Martynas Mankus and Matan Hasson and Matthew Lennie and Matthias Reso and Maxim Groshev and Maxim Naumov and Maya Lathi and Meghan Keneally and Miao Liu and Michael L. Seltzer and Michal Valko and Michelle Restrepo and Mihir Patel and Mik Vyatskov and Mikayel Samvelyan and Mike Clark and Mike Macey and Mike Wang and Miquel Jubert Hermoso and Mo Metanat and Mohammad Rastegari and Munish Bansal and Nandhini Santhanam and Natascha Parks and Natasha White and Navyata Bawa and Nayan Singhal and Nick Egebo and Nicolas Usunier and Nikhil Mehta and Nikolay Pavlovich Laptev and Ning Dong and Norman Cheng and Oleg Chernoguz and Olivia Hart and Omkar Salpekar and Ozlem Kalinli and Parkin Kent and Parth Parekh and Paul Saab and Pavan Balaji and Pedro Rittner and Philip Bontrager and Pierre Roux and Piotr Dollar and Polina Zvyagina and Prashant Ratanchandani and Pritish Yuvraj and Qian Liang and Rachad Alao and Rachel Rodriguez and Rafi Ayub and Raghotham Murthy and Raghu Nayani and Rahul Mitra and Rangaprabhu Parthasarathy and Raymond Li and Rebekkah Hogan and Robin Battey and Rocky Wang and Russ Howes and Ruty Rinott and Sachin Mehta and Sachin Siby and Sai Jayesh Bondu and Samyak Datta and Sara Chugh and Sara Hunt and Sargun Dhillon and Sasha Sidorov and Satadru Pan and Saurabh Mahajan and Saurabh Verma and Seiji Yamamoto and Sharadh Ramaswamy and Shaun Lindsay and Shaun Lindsay and Sheng Feng and Shenghao Lin and Shengxin Cindy Zha and Shishir Patil and Shiva Shankar and Shuqiang Zhang and Shuqiang Zhang and Sinong Wang and Sneha Agarwal and Soji Sajuyigbe and Soumith Chintala and Stephanie Max and Stephen Chen and Steve Kehoe and Steve Satterfield and Sudarshan Govindaprasad and Sumit Gupta and Summer Deng and Sungmin Cho and Sunny Virk and Suraj Subramanian and Sy Choudhury and Sydney Goldman and Tal Remez and Tamar Glaser and Tamara Best and Thilo Koehler and Thomas Robinson and Tianhe Li and Tianjun Zhang and Tim Matthews and Timothy Chou and Tzook Shaked and Varun Vontimitta and Victoria Ajayi and Victoria Montanez and Vijai Mohan and Vinay Satish Kumar and Vishal Mangla and Vlad Ionescu and Vlad Poenaru and Vlad Tiberiu Mihailescu and Vladimir Ivanov and Wei Li and Wenchen Wang and Wenwen Jiang and Wes Bouaziz and Will Constable and Xiaocheng Tang and Xiaojian Wu and Xiaolan Wang and Xilun Wu and Xinbo Gao and Yaniv Kleinman and Yanjun Chen and Ye Hu and Ye Jia and Ye Qi and Yenda Li and Yilin Zhang and Ying Zhang and Yossi Adi and Youngjin Nam and Yu and Wang and Yu Zhao and Yuchen Hao and Yundi Qian and Yunlu Li and Yuzi He and Zach Rait and Zachary DeVito and Zef Rosnbrick and Zhaoduo Wen and Zhenyu Yang and Zhiwei Zhao and Zhiyu Ma},
      year={2024},
      eprint={2407.21783},
      archivePrefix={arXiv},
      primaryClass={cs.AI},
      url={https://arxiv.org/abs/2407.21783}, 
}

@misc{yang2025qwen3technicalreport,
      title={Qwen3 Technical Report}, 
      author={An Yang and Anfeng Li and Baosong Yang and Beichen Zhang and Binyuan Hui and Bo Zheng and Bowen Yu and Chang Gao and Chengen Huang and Chenxu Lv and Chujie Zheng and Dayiheng Liu and Fan Zhou and Fei Huang and Feng Hu and Hao Ge and Haoran Wei and Huan Lin and Jialong Tang and Jian Yang and Jianhong Tu and Jianwei Zhang and Jianxin Yang and Jiaxi Yang and Jing Zhou and Jingren Zhou and Junyang Lin and Kai Dang and Keqin Bao and Kexin Yang and Le Yu and Lianghao Deng and Mei Li and Mingfeng Xue and Mingze Li and Pei Zhang and Peng Wang and Qin Zhu and Rui Men and Ruize Gao and Shixuan Liu and Shuang Luo and Tianhao Li and Tianyi Tang and Wenbiao Yin and Xingzhang Ren and Xinyu Wang and Xinyu Zhang and Xuancheng Ren and Yang Fan and Yang Su and Yichang Zhang and Yinger Zhang and Yu Wan and Yuqiong Liu and Zekun Wang and Zeyu Cui and Zhenru Zhang and Zhipeng Zhou and Zihan Qiu},
      year={2025},
      eprint={2505.09388},
      archivePrefix={arXiv},
      primaryClass={cs.CL},
      url={https://arxiv.org/abs/2505.09388}, 
}

@misc{deepseekai2025deepseekv3technicalreport,
      title={DeepSeek-V3 Technical Report}, 
      author={DeepSeek-AI and Aixin Liu and Bei Feng and Bing Xue and Bingxuan Wang and Bochao Wu and Chengda Lu and Chenggang Zhao and Chengqi Deng and Chenyu Zhang and Chong Ruan and Damai Dai and Daya Guo and Dejian Yang and Deli Chen and Dongjie Ji and Erhang Li and Fangyun Lin and Fucong Dai and Fuli Luo and Guangbo Hao and Guanting Chen and Guowei Li and H. Zhang and Han Bao and Hanwei Xu and Haocheng Wang and Haowei Zhang and Honghui Ding and Huajian Xin and Huazuo Gao and Hui Li and Hui Qu and J. L. Cai and Jian Liang and Jianzhong Guo and Jiaqi Ni and Jiashi Li and Jiawei Wang and Jin Chen and Jingchang Chen and Jingyang Yuan and Junjie Qiu and Junlong Li and Junxiao Song and Kai Dong and Kai Hu and Kaige Gao and Kang Guan and Kexin Huang and Kuai Yu and Lean Wang and Lecong Zhang and Lei Xu and Leyi Xia and Liang Zhao and Litong Wang and Liyue Zhang and Meng Li and Miaojun Wang and Mingchuan Zhang and Minghua Zhang and Minghui Tang and Mingming Li and Ning Tian and Panpan Huang and Peiyi Wang and Peng Zhang and Qiancheng Wang and Qihao Zhu and Qinyu Chen and Qiushi Du and R. J. Chen and R. L. Jin and Ruiqi Ge and Ruisong Zhang and Ruizhe Pan and Runji Wang and Runxin Xu and Ruoyu Zhang and Ruyi Chen and S. S. Li and Shanghao Lu and Shangyan Zhou and Shanhuang Chen and Shaoqing Wu and Shengfeng Ye and Shengfeng Ye and Shirong Ma and Shiyu Wang and Shuang Zhou and Shuiping Yu and Shunfeng Zhou and Shuting Pan and T. Wang and Tao Yun and Tian Pei and Tianyu Sun and W. L. Xiao and Wangding Zeng and Wanjia Zhao and Wei An and Wen Liu and Wenfeng Liang and Wenjun Gao and Wenqin Yu and Wentao Zhang and X. Q. Li and Xiangyue Jin and Xianzu Wang and Xiao Bi and Xiaodong Liu and Xiaohan Wang and Xiaojin Shen and Xiaokang Chen and Xiaokang Zhang and Xiaosha Chen and Xiaotao Nie and Xiaowen Sun and Xiaoxiang Wang and Xin Cheng and Xin Liu and Xin Xie and Xingchao Liu and Xingkai Yu and Xinnan Song and Xinxia Shan and Xinyi Zhou and Xinyu Yang and Xinyuan Li and Xuecheng Su and Xuheng Lin and Y. K. Li and Y. Q. Wang and Y. X. Wei and Y. X. Zhu and Yang Zhang and Yanhong Xu and Yanhong Xu and Yanping Huang and Yao Li and Yao Zhao and Yaofeng Sun and Yaohui Li and Yaohui Wang and Yi Yu and Yi Zheng and Yichao Zhang and Yifan Shi and Yiliang Xiong and Ying He and Ying Tang and Yishi Piao and Yisong Wang and Yixuan Tan and Yiyang Ma and Yiyuan Liu and Yongqiang Guo and Yu Wu and Yuan Ou and Yuchen Zhu and Yuduan Wang and Yue Gong and Yuheng Zou and Yujia He and Yukun Zha and Yunfan Xiong and Yunxian Ma and Yuting Yan and Yuxiang Luo and Yuxiang You and Yuxuan Liu and Yuyang Zhou and Z. F. Wu and Z. Z. Ren and Zehui Ren and Zhangli Sha and Zhe Fu and Zhean Xu and Zhen Huang and Zhen Zhang and Zhenda Xie and Zhengyan Zhang and Zhewen Hao and Zhibin Gou and Zhicheng Ma and Zhigang Yan and Zhihong Shao and Zhipeng Xu and Zhiyu Wu and Zhongyu Zhang and Zhuoshu Li and Zihui Gu and Zijia Zhu and Zijun Liu and Zilin Li and Ziwei Xie and Ziyang Song and Ziyi Gao and Zizheng Pan},
      year={2025},
      eprint={2412.19437},
      archivePrefix={arXiv},
      primaryClass={cs.CL},
      url={https://arxiv.org/abs/2412.19437}, 
}

@misc{gemmateam2025gemma3technicalreport,
      title={Gemma 3 Technical Report}, 
      author={Gemma Team and Aishwarya Kamath and Johan Ferret and Shreya Pathak and Nino Vieillard and Ramona Merhej and Sarah Perrin and Tatiana Matejovicova and Alexandre Ramé and Morgane Rivière and Louis Rouillard and Thomas Mesnard and Geoffrey Cideron and Jean-bastien Grill and Sabela Ramos and Edouard Yvinec and Michelle Casbon and Etienne Pot and Ivo Penchev and Gaël Liu and Francesco Visin and Kathleen Kenealy and Lucas Beyer and Xiaohai Zhai and Anton Tsitsulin and Robert Busa-Fekete and Alex Feng and Noveen Sachdeva and Benjamin Coleman and Yi Gao and Basil Mustafa and Iain Barr and Emilio Parisotto and David Tian and Matan Eyal and Colin Cherry and Jan-Thorsten Peter and Danila Sinopalnikov and Surya Bhupatiraju and Rishabh Agarwal and Mehran Kazemi and Dan Malkin and Ravin Kumar and David Vilar and Idan Brusilovsky and Jiaming Luo and Andreas Steiner and Abe Friesen and Abhanshu Sharma and Abheesht Sharma and Adi Mayrav Gilady and Adrian Goedeckemeyer and Alaa Saade and Alex Feng and Alexander Kolesnikov and Alexei Bendebury and Alvin Abdagic and Amit Vadi and András György and André Susano Pinto and Anil Das and Ankur Bapna and Antoine Miech and Antoine Yang and Antonia Paterson and Ashish Shenoy and Ayan Chakrabarti and Bilal Piot and Bo Wu and Bobak Shahriari and Bryce Petrini and Charlie Chen and Charline Le Lan and Christopher A. Choquette-Choo and CJ Carey and Cormac Brick and Daniel Deutsch and Danielle Eisenbud and Dee Cattle and Derek Cheng and Dimitris Paparas and Divyashree Shivakumar Sreepathihalli and Doug Reid and Dustin Tran and Dustin Zelle and Eric Noland and Erwin Huizenga and Eugene Kharitonov and Frederick Liu and Gagik Amirkhanyan and Glenn Cameron and Hadi Hashemi and Hanna Klimczak-Plucińska and Harman Singh and Harsh Mehta and Harshal Tushar Lehri and Hussein Hazimeh and Ian Ballantyne and Idan Szpektor and Ivan Nardini and Jean Pouget-Abadie and Jetha Chan and Joe Stanton and John Wieting and Jonathan Lai and Jordi Orbay and Joseph Fernandez and Josh Newlan and Ju-yeong Ji and Jyotinder Singh and Kat Black and Kathy Yu and Kevin Hui and Kiran Vodrahalli and Klaus Greff and Linhai Qiu and Marcella Valentine and Marina Coelho and Marvin Ritter and Matt Hoffman and Matthew Watson and Mayank Chaturvedi and Michael Moynihan and Min Ma and Nabila Babar and Natasha Noy and Nathan Byrd and Nick Roy and Nikola Momchev and Nilay Chauhan and Noveen Sachdeva and Oskar Bunyan and Pankil Botarda and Paul Caron and Paul Kishan Rubenstein and Phil Culliton and Philipp Schmid and Pier Giuseppe Sessa and Pingmei Xu and Piotr Stanczyk and Pouya Tafti and Rakesh Shivanna and Renjie Wu and Renke Pan and Reza Rokni and Rob Willoughby and Rohith Vallu and Ryan Mullins and Sammy Jerome and Sara Smoot and Sertan Girgin and Shariq Iqbal and Shashir Reddy and Shruti Sheth and Siim Põder and Sijal Bhatnagar and Sindhu Raghuram Panyam and Sivan Eiger and Susan Zhang and Tianqi Liu and Trevor Yacovone and Tyler Liechty and Uday Kalra and Utku Evci and Vedant Misra and Vincent Roseberry and Vlad Feinberg and Vlad Kolesnikov and Woohyun Han and Woosuk Kwon and Xi Chen and Yinlam Chow and Yuvein Zhu and Zichuan Wei and Zoltan Egyed and Victor Cotruta and Minh Giang and Phoebe Kirk and Anand Rao and Kat Black and Nabila Babar and Jessica Lo and Erica Moreira and Luiz Gustavo Martins and Omar Sanseviero and Lucas Gonzalez and Zach Gleicher and Tris Warkentin and Vahab Mirrokni and Evan Senter and Eli Collins and Joelle Barral and Zoubin Ghahramani and Raia Hadsell and Yossi Matias and D. Sculley and Slav Petrov and Noah Fiedel and Noam Shazeer and Oriol Vinyals and Jeff Dean and Demis Hassabis and Koray Kavukcuoglu and Clement Farabet and Elena Buchatskaya and Jean-Baptiste Alayrac and Rohan Anil and Dmitry and Lepikhin and Sebastian Borgeaud and Olivier Bachem and Armand Joulin and Alek Andreev and Cassidy Hardin and Robert Dadashi and Léonard Hussenot},
      year={2025},
      eprint={2503.19786},
      archivePrefix={arXiv},
      primaryClass={cs.CL},
      url={https://arxiv.org/abs/2503.19786}, 
}

@inproceedings{zaratiana-etal-2024-gliner,
    title = "{GL}i{NER}: Generalist Model for Named Entity Recognition using Bidirectional Transformer",
    author = "Zaratiana, Urchade  and
      Tomeh, Nadi  and
      Holat, Pierre  and
      Charnois, Thierry",
    editor = "Duh, Kevin  and
      Gomez, Helena  and
      Bethard, Steven",
    booktitle = "Proceedings of the 2024 Conference of the North American Chapter of the Association for Computational Linguistics: Human Language Technologies (Volume 1: Long Papers)",
    month = jun,
    year = "2024",
    address = "Mexico City, Mexico",
    publisher = "Association for Computational Linguistics",
    url = "https://aclanthology.org/2024.naacl-long.300/",
    doi = "10.18653/v1/2024.naacl-long.300",
    pages = "5364--5376"
}

@inproceedings{zaratiana-etal-2025-gliner2,
    title = "{GL}i{NER}2: Schema-Driven Multi-Task Learning for Structured Information Extraction",
    author = "Zaratiana, Urchade  and
      Pasternak, Gil  and
      Boyd, Oliver  and
      Hurn-Maloney, George  and
      Lewis, Ash",
    editor = {Habernal, Ivan  and
      Schulam, Peter  and
      Tiedemann, J{\"o}rg},
    booktitle = "Proceedings of the 2025 Conference on Empirical Methods in Natural Language Processing: System Demonstrations",
    month = nov,
    year = "2025",
    address = "Suzhou, China",
    publisher = "Association for Computational Linguistics",
    url = "https://aclanthology.org/2025.emnlp-demos.10/",
    doi = "10.18653/v1/2025.emnlp-demos.10",
    pages = "130--140",
    ISBN = "979-8-89176-334-0"
}

@inproceedings{bogdanov-etal-2024-nuner,
    title = "{N}u{NER}: Entity Recognition Encoder Pre-training via {LLM}-Annotated Data",
    author = "Bogdanov, Sergei  and
      Constantin, Alexandre  and
      Bernard, Timoth{\'e}e  and
      Crabb{\'e}, Benoit  and
      Bernard, Etienne P",
    editor = "Al-Onaizan, Yaser  and
      Bansal, Mohit  and
      Chen, Yun-Nung",
    booktitle = "Proceedings of the 2024 Conference on Empirical Methods in Natural Language Processing",
    month = nov,
    year = "2024",
    address = "Miami, Florida, USA",
    publisher = "Association for Computational Linguistics",
    url = "https://aclanthology.org/2024.emnlp-main.660/",
    doi = "10.18653/v1/2024.emnlp-main.660",
    pages = "11829--11841"
}

@inproceedings{golde-etal-2024-large,
    title = "Large-Scale Label Interpretation Learning for Few-Shot Named Entity Recognition",
    author = "Golde, Jonas  and
      Hamborg, Felix  and
      Akbik, Alan",
    editor = "Graham, Yvette  and
      Purver, Matthew",
    booktitle = "Proceedings of the 18th Conference of the European Chapter of the Association for Computational Linguistics (Volume 1: Long Papers)",
    month = mar,
    year = "2024",
    address = "St. Julian{'}s, Malta",
    publisher = "Association for Computational Linguistics",
    url = "https://aclanthology.org/2024.eacl-long.178/",
    doi = "10.18653/v1/2024.eacl-long.178",
    pages = "2915--2930"
}

@inproceedings{
    zhang2023optimizing,
    title={Optimizing Bi-Encoder for Named Entity Recognition via Contrastive Learning},
    author={Sheng Zhang and Hao Cheng and Jianfeng Gao and Hoifung Poon},
    booktitle={The Eleventh International Conference on Learning Representations },
    year={2023},
    url={https://openreview.net/forum?id=9EAQVEINuum}
}

@misc{he2023debertav3improvingdebertausing,
      title={DeBERTaV3: Improving DeBERTa using ELECTRA-Style Pre-Training with Gradient-Disentangled Embedding Sharing}, 
      author={Pengcheng He and Jianfeng Gao and Weizhu Chen},
      year={2023},
      eprint={2111.09543},
      archivePrefix={arXiv},
      primaryClass={cs.CL},
      url={https://arxiv.org/abs/2111.09543}, 
}

@inproceedings{xue-etal-2021-mt5,
    title = "m{T}5: A Massively Multilingual Pre-trained Text-to-Text Transformer",
    author = "Xue, Linting  and
      Constant, Noah  and
      Roberts, Adam  and
      Kale, Mihir  and
      Al-Rfou, Rami  and
      Siddhant, Aditya  and
      Barua, Aditya  and
      Raffel, Colin",
    editor = "Toutanova, Kristina  and
      Rumshisky, Anna  and
      Zettlemoyer, Luke  and
      Hakkani-Tur, Dilek  and
      Beltagy, Iz  and
      Bethard, Steven  and
      Cotterell, Ryan  and
      Chakraborty, Tanmoy  and
      Zhou, Yichao",
    booktitle = "Proceedings of the 2021 Conference of the North American Chapter of the Association for Computational Linguistics: Human Language Technologies",
    month = jun,
    year = "2021",
    address = "Online",
    publisher = "Association for Computational Linguistics",
    url = "https://aclanthology.org/2021.naacl-main.41/",
    doi = "10.18653/v1/2021.naacl-main.41",
    pages = "483--498"
}

@misc{brown2020languagemodelsfewshotlearners,
      title={Language Models are Few-Shot Learners}, 
      author={Tom B. Brown and Benjamin Mann and Nick Ryder and Melanie Subbiah and Jared Kaplan and Prafulla Dhariwal and Arvind Neelakantan and Pranav Shyam and Girish Sastry and Amanda Askell and Sandhini Agarwal and Ariel Herbert-Voss and Gretchen Krueger and Tom Henighan and Rewon Child and Aditya Ramesh and Daniel M. Ziegler and Jeffrey Wu and Clemens Winter and Christopher Hesse and Mark Chen and Eric Sigler and Mateusz Litwin and Scott Gray and Benjamin Chess and Jack Clark and Christopher Berner and Sam McCandlish and Alec Radford and Ilya Sutskever and Dario Amodei},
      year={2020},
      eprint={2005.14165},
      archivePrefix={arXiv},
      primaryClass={cs.CL},
      url={https://arxiv.org/abs/2005.14165}, 
}

@inproceedings{schick-schutze-2021-just,
    title = "It{'}s Not Just Size That Matters: Small Language Models Are Also Few-Shot Learners",
    author = {Schick, Timo  and
      Sch{\"u}tze, Hinrich},
    editor = "Toutanova, Kristina  and
      Rumshisky, Anna  and
      Zettlemoyer, Luke  and
      Hakkani-Tur, Dilek  and
      Beltagy, Iz  and
      Bethard, Steven  and
      Cotterell, Ryan  and
      Chakraborty, Tanmoy  and
      Zhou, Yichao",
    booktitle = "Proceedings of the 2021 Conference of the North American Chapter of the Association for Computational Linguistics: Human Language Technologies",
    month = jun,
    year = "2021",
    address = "Online",
    publisher = "Association for Computational Linguistics",
    url = "https://aclanthology.org/2021.naacl-main.185/",
    doi = "10.18653/v1/2021.naacl-main.185",
    pages = "2339--2352"
}

@inproceedings{min-etal-2022-rethinking,
    title = "Rethinking the Role of Demonstrations: What Makes In-Context Learning Work?",
    author = "Min, Sewon  and
      Lyu, Xinxi  and
      Holtzman, Ari  and
      Artetxe, Mikel  and
      Lewis, Mike  and
      Hajishirzi, Hannaneh  and
      Zettlemoyer, Luke",
    editor = "Goldberg, Yoav  and
      Kozareva, Zornitsa  and
      Zhang, Yue",
    booktitle = "Proceedings of the 2022 Conference on Empirical Methods in Natural Language Processing",
    month = dec,
    year = "2022",
    address = "Abu Dhabi, United Arab Emirates",
    publisher = "Association for Computational Linguistics",
    url = "https://aclanthology.org/2022.emnlp-main.759/",
    doi = "10.18653/v1/2022.emnlp-main.759",
    pages = "11048--11064"
}

@misc{lou2023universalinformationextractionunified,
      title={Universal Information Extraction as Unified Semantic Matching}, 
      author={Jie Lou and Yaojie Lu and Dai Dai and Wei Jia and Hongyu Lin and Xianpei Han and Le Sun and Hua Wu},
      year={2023},
      eprint={2301.03282},
      archivePrefix={arXiv},
      primaryClass={cs.CL},
      url={https://arxiv.org/abs/2301.03282}, 
}

@misc{wang2023instructuiemultitaskinstructiontuning,
      title={InstructUIE: Multi-task Instruction Tuning for Unified Information Extraction}, 
      author={Xiao Wang and Weikang Zhou and Can Zu and Han Xia and Tianze Chen and Yuansen Zhang and Rui Zheng and Junjie Ye and Qi Zhang and Tao Gui and Jihua Kang and Jingsheng Yang and Siyuan Li and Chunsai Du},
      year={2023},
      eprint={2304.08085},
      archivePrefix={arXiv},
      primaryClass={cs.CL},
      url={https://arxiv.org/abs/2304.08085}, 
}

@misc{zhou2024universalnertargeteddistillationlarge,
      title={UniversalNER: Targeted Distillation from Large Language Models for Open Named Entity Recognition}, 
      author={Wenxuan Zhou and Sheng Zhang and Yu Gu and Muhao Chen and Hoifung Poon},
      year={2024},
      eprint={2308.03279},
      archivePrefix={arXiv},
      primaryClass={cs.CL},
      url={https://arxiv.org/abs/2308.03279}, 
}

@inproceedings{sun-etal-2023-text,
    title = "Text Classification via Large Language Models",
    author = "Sun, Xiaofei  and
      Li, Xiaoya  and
      Li, Jiwei  and
      Wu, Fei  and
      Guo, Shangwei  and
      Zhang, Tianwei  and
      Wang, Guoyin",
    editor = "Bouamor, Houda  and
      Pino, Juan  and
      Bali, Kalika",
    booktitle = "Findings of the Association for Computational Linguistics: EMNLP 2023",
    month = dec,
    year = "2023",
    address = "Singapore",
    publisher = "Association for Computational Linguistics",
    url = "https://aclanthology.org/2023.findings-emnlp.603/",
    doi = "10.18653/v1/2023.findings-emnlp.603",
    pages = "8990--9005"
}

@misc{decao2021autoregressiveentityretrieval,
      title={Autoregressive Entity Retrieval}, 
      author={Nicola De Cao and Gautier Izacard and Sebastian Riedel and Fabio Petroni},
      year={2021},
      eprint={2010.00904},
      archivePrefix={arXiv},
      primaryClass={cs.CL},
      url={https://arxiv.org/abs/2010.00904}, 
}

@inproceedings{ding-etal-2024-chatel,
    title = "{C}hat{EL}: Entity Linking with Chatbots",
    author = "Ding, Yifan  and
      Zeng, Qingkai  and
      Weninger, Tim",
    editor = "Calzolari, Nicoletta  and
      Kan, Min-Yen  and
      Hoste, Veronique  and
      Lenci, Alessandro  and
      Sakti, Sakriani  and
      Xue, Nianwen",
    booktitle = "Proceedings of the 2024 Joint International Conference on Computational Linguistics, Language Resources and Evaluation (LREC-COLING 2024)",
    month = may,
    year = "2024",
    address = "Torino, Italia",
    publisher = "ELRA and ICCL",
    url = "https://aclanthology.org/2024.lrec-main.275/",
    pages = "3086--3097"
}

@misc{ashok2023promptnerpromptingnamedentity,
      title={PromptNER: Prompting For Named Entity Recognition}, 
      author={Dhananjay Ashok and Zachary C. Lipton},
      year={2023},
      eprint={2305.15444},
      archivePrefix={arXiv},
      primaryClass={cs.CL},
      url={https://arxiv.org/abs/2305.15444}, 
}

@inproceedings{huang-etal-2022-copner,
    title = "{COPNER}: Contrastive Learning with Prompt Guiding for Few-shot Named Entity Recognition",
    author = "Huang, Yucheng  and
      He, Kai  and
      Wang, Yige  and
      Zhang, Xianli  and
      Gong, Tieliang  and
      Mao, Rui  and
      Li, Chen",
    editor = "Calzolari, Nicoletta  and
      Huang, Chu-Ren  and
      Kim, Hansaem  and
      Pustejovsky, James  and
      Wanner, Leo  and
      Choi, Key-Sun  and
      Ryu, Pum-Mo  and
      Chen, Hsin-Hsi  and
      Donatelli, Lucia  and
      Ji, Heng  and
      Kurohashi, Sadao  and
      Paggio, Patrizia  and
      Xue, Nianwen  and
      Kim, Seokhwan  and
      Hahm, Younggyun  and
      He, Zhong  and
      Lee, Tony Kyungil  and
      Santus, Enrico  and
      Bond, Francis  and
      Na, Seung-Hoon",
    booktitle = "Proceedings of the 29th International Conference on Computational Linguistics",
    month = oct,
    year = "2022",
    address = "Gyeongju, Republic of Korea",
    publisher = "International Committee on Computational Linguistics",
    url = "https://aclanthology.org/2022.coling-1.222/",
    pages = "2515--2527"
}

@inproceedings{halder-etal-2020-task,
    title = "Task-Aware Representation of Sentences for Generic Text Classification",
    author = "Halder, Kishaloy  and
      Akbik, Alan  and
      Krapac, Josip  and
      Vollgraf, Roland",
    editor = "Scott, Donia  and
      Bel, Nuria  and
      Zong, Chengqing",
    booktitle = "Proceedings of the 28th International Conference on Computational Linguistics",
    month = dec,
    year = "2020",
    address = "Barcelona, Spain (Online)",
    publisher = "International Committee on Computational Linguistics",
    url = "https://aclanthology.org/2020.coling-main.285/",
    doi = "10.18653/v1/2020.coling-main.285",
    pages = "3202--3213"
}

@misc{golde2025finerwebdatasetsartifactsscalable,
      title={FiNERweb: Datasets and Artifacts for Scalable Multilingual Named Entity Recognition}, 
      author={Jonas Golde and Patrick Haller and Alan Akbik},
      year={2025},
      eprint={2512.13884},
      archivePrefix={arXiv},
      primaryClass={cs.CL},
      url={https://arxiv.org/abs/2512.13884}, 
}

@misc{hinton2015distillingknowledgeneuralnetwork,
      title={Distilling the Knowledge in a Neural Network}, 
      author={Geoffrey Hinton and Oriol Vinyals and Jeff Dean},
      year={2015},
      eprint={1503.02531},
      archivePrefix={arXiv},
      primaryClass={stat.ML},
      url={https://arxiv.org/abs/1503.02531}, 
}

@inproceedings{ye-etal-2022-zerogen,
    title = "{Z}ero{G}en: Efficient Zero-shot Learning via Dataset Generation",
    author = "Ye, Jiacheng  and
      Gao, Jiahui  and
      Li, Qintong  and
      Xu, Hang  and
      Feng, Jiangtao  and
      Wu, Zhiyong  and
      Yu, Tao  and
      Kong, Lingpeng",
    editor = "Goldberg, Yoav  and
      Kozareva, Zornitsa  and
      Zhang, Yue",
    booktitle = "Proceedings of the 2022 Conference on Empirical Methods in Natural Language Processing",
    month = dec,
    year = "2022",
    address = "Abu Dhabi, United Arab Emirates",
    publisher = "Association for Computational Linguistics",
    url = "https://aclanthology.org/2022.emnlp-main.801/",
    doi = "10.18653/v1/2022.emnlp-main.801",
    pages = "11653--11669"
}

@inproceedings{huang-etal-2019-matters,
    title = "What Matters for Neural Cross-Lingual Named Entity Recognition: An Empirical Analysis",
    author = "Huang, Xiaolei  and
      May, Jonathan  and
      Peng, Nanyun",
    editor = "Inui, Kentaro  and
      Jiang, Jing  and
      Ng, Vincent  and
      Wan, Xiaojun",
    booktitle = "Proceedings of the 2019 Conference on Empirical Methods in Natural Language Processing and the 9th International Joint Conference on Natural Language Processing (EMNLP-IJCNLP)",
    month = nov,
    year = "2019",
    address = "Hong Kong, China",
    publisher = "Association for Computational Linguistics",
    url = "https://aclanthology.org/D19-1672/",
    doi = "10.18653/v1/D19-1672",
    pages = "6395--6401"
}

@inproceedings{devlin-etal-2019-bert,
    title = "{BERT}: Pre-training of Deep Bidirectional Transformers for Language Understanding",
    author = "Devlin, Jacob  and
      Chang, Ming-Wei  and
      Lee, Kenton  and
      Toutanova, Kristina",
    editor = "Burstein, Jill  and
      Doran, Christy  and
      Solorio, Thamar",
    booktitle = "Proceedings of the 2019 Conference of the North {A}merican Chapter of the Association for Computational Linguistics: Human Language Technologies, Volume 1 (Long and Short Papers)",
    month = jun,
    year = "2019",
    address = "Minneapolis, Minnesota",
    publisher = "Association for Computational Linguistics",
    url = "https://aclanthology.org/N19-1423/",
    doi = "10.18653/v1/N19-1423",
    pages = "4171--4186"
}

@misc{loshchilov2019decoupledweightdecayregularization,
      title={Decoupled Weight Decay Regularization}, 
      author={Ilya Loshchilov and Frank Hutter},
      year={2019},
      eprint={1711.05101},
      archivePrefix={arXiv},
      primaryClass={cs.LG},
      url={https://arxiv.org/abs/1711.05101}, 
}

@inproceedings{luo-etal-2025-dynamicner,
    title = "{D}ynamic{NER}: A Dynamic, Multilingual, and Fine-Grained Dataset for {LLM}-based Named Entity Recognition",
    author = "Luo, Hanjun  and
      Jin, Yingbin  and
      Wang, Yiran  and
      Li, Xinfeng  and
      Shang, Tong  and
      Liu, Xuecheng  and
      Chen, Ruizhe  and
      Wang, Kun  and
      Salam, Hanan  and
      Wen, Qingsong  and
      Liu, Zuozhu",
    editor = "Christodoulopoulos, Christos  and
      Chakraborty, Tanmoy  and
      Rose, Carolyn  and
      Peng, Violet",
    booktitle = "Proceedings of the 2025 Conference on Empirical Methods in Natural Language Processing",
    month = nov,
    year = "2025",
    address = "Suzhou, China",
    publisher = "Association for Computational Linguistics",
    url = "https://aclanthology.org/2025.emnlp-main.835/",
    doi = "10.18653/v1/2025.emnlp-main.835",
    pages = "16522--16546",
    ISBN = "979-8-89176-332-6"
}

@inproceedings{mayhew-etal-2024-universal,
    title = "Universal {NER}: A Gold-Standard Multilingual Named Entity Recognition Benchmark",
    author = {Mayhew, Stephen  and
      Blevins, Terra  and
      Liu, Shuheng  and
      {\v{S}}uppa, Marek  and
      Gonen, Hila  and
      Imperial, Joseph Marvin  and
      Karlsson, B{\"o}rje F.  and
      Lin, Peiqin  and
      Ljube{\v{s}}i{\'c}, Nikola  and
      Miranda, LJ  and
      Plank, Barbara  and
      Riabi, Arij  and
      Pinter, Yuval},
    editor = "Duh, Kevin  and
      Gomez, Helena  and
      Bethard, Steven",
    booktitle = "Proceedings of the 2024 Conference of the North American Chapter of the Association for Computational Linguistics: Human Language Technologies (Volume 1: Long Papers)",
    month = jun,
    year = "2024",
    address = "Mexico City, Mexico",
    publisher = "Association for Computational Linguistics",
    url = "https://aclanthology.org/2024.naacl-long.243/",
    doi = "10.18653/v1/2024.naacl-long.243",
    pages = "4322--4337"
}

@inproceedings{adelani-etal-2022-masakhaner,
    title = "{M}asakha{NER} 2.0: {A}frica-centric Transfer Learning for Named Entity Recognition",
    author = "Adelani, David Ifeoluwa  and
      Neubig, Graham  and
      Ruder, Sebastian  and
      Rijhwani, Shruti  and
      Beukman, Michael  and
      Palen-Michel, Chester  and
      Lignos, Constantine  and
      Alabi, Jesujoba O.  and
      Muhammad, Shamsuddeen H.  and
      Nabende, Peter  and
      Dione, Cheikh M. Bamba  and
      Bukula, Andiswa  and
      Mabuya, Rooweither  and
      Dossou, Bonaventure F. P.  and
      Sibanda, Blessing  and
      Buzaaba, Happy  and
      Mukiibi, Jonathan  and
      Kalipe, Godson  and
      Mbaye, Derguene  and
      Taylor, Amelia  and
      Kabore, Fatoumata  and
      Emezue, Chris Chinenye  and
      Aremu, Anuoluwapo  and
      Ogayo, Perez  and
      Gitau, Catherine  and
      Munkoh-Buabeng, Edwin  and
      Memdjokam Koagne, Victoire  and
      Tapo, Allahsera Auguste  and
      Macucwa, Tebogo  and
      Marivate, Vukosi  and
      Mboning, Elvis  and
      Gwadabe, Tajuddeen  and
      Adewumi, Tosin  and
      Ahia, Orevaoghene  and
      Nakatumba-Nabende, Joyce  and
      Mokono, Neo L.  and
      Ezeani, Ignatius  and
      Chukwuneke, Chiamaka  and
      Adeyemi, Mofetoluwa  and
      Hacheme, Gilles Q.  and
      Abdulmumin, Idris  and
      Ogundepo, Odunayo  and
      Yousuf, Oreen  and
      Moteu Ngoli, Tatiana  and
      Klakow, Dietrich",
    editor = "Goldberg, Yoav  and
      Kozareva, Zornitsa  and
      Zhang, Yue",
    booktitle = "Proceedings of the 2022 Conference on Empirical Methods in Natural Language Processing",
    month = dec,
    year = "2022",
    address = "Abu Dhabi, United Arab Emirates",
    publisher = "Association for Computational Linguistics",
    url = "https://aclanthology.org/2022.emnlp-main.298/",
    doi = "10.18653/v1/2022.emnlp-main.298",
    pages = "4488--4508"
}

@inproceedings{tedeschi-navigli-2022-multinerd,
    title = "{M}ulti{NERD}: A Multilingual, Multi-Genre and Fine-Grained Dataset for Named Entity Recognition (and Disambiguation)",
    author = "Tedeschi, Simone  and
      Navigli, Roberto",
    editor = "Carpuat, Marine  and
      de Marneffe, Marie-Catherine  and
      Meza Ruiz, Ivan Vladimir",
    booktitle = "Findings of the Association for Computational Linguistics: NAACL 2022",
    month = jul,
    year = "2022",
    address = "Seattle, United States",
    publisher = "Association for Computational Linguistics",
    url = "https://aclanthology.org/2022.findings-naacl.60/",
    doi = "10.18653/v1/2022.findings-naacl.60",
    pages = "801--812"
}

@inproceedings{malmasi-etal-2022-multiconer,
    title = "{M}ulti{C}o{NER}: A Large-scale Multilingual Dataset for Complex Named Entity Recognition",
    author = "Malmasi, Shervin  and
      Fang, Anjie  and
      Fetahu, Besnik  and
      Kar, Sudipta  and
      Rokhlenko, Oleg",
    editor = "Calzolari, Nicoletta  and
      Huang, Chu-Ren  and
      Kim, Hansaem  and
      Pustejovsky, James  and
      Wanner, Leo  and
      Choi, Key-Sun  and
      Ryu, Pum-Mo  and
      Chen, Hsin-Hsi  and
      Donatelli, Lucia  and
      Ji, Heng  and
      Kurohashi, Sadao  and
      Paggio, Patrizia  and
      Xue, Nianwen  and
      Kim, Seokhwan  and
      Hahm, Younggyun  and
      He, Zhong  and
      Lee, Tony Kyungil  and
      Santus, Enrico  and
      Bond, Francis  and
      Na, Seung-Hoon",
    booktitle = "Proceedings of the 29th International Conference on Computational Linguistics",
    month = oct,
    year = "2022",
    address = "Gyeongju, Republic of Korea",
    publisher = "International Committee on Computational Linguistics",
    url = "https://aclanthology.org/2022.coling-1.334/",
    pages = "3798--3809"
}

@inproceedings{fetahu-etal-2023-multiconer,
    title = "{M}ulti{C}o{NER} v2: a Large Multilingual dataset for Fine-grained and Noisy Named Entity Recognition",
    author = "Fetahu, Besnik  and
      Chen, Zhiyu  and
      Kar, Sudipta  and
      Rokhlenko, Oleg  and
      Malmasi, Shervin",
    editor = "Bouamor, Houda  and
      Pino, Juan  and
      Bali, Kalika",
    booktitle = "Findings of the Association for Computational Linguistics: EMNLP 2023",
    month = dec,
    year = "2023",
    address = "Singapore",
    publisher = "Association for Computational Linguistics",
    url = "https://aclanthology.org/2023.findings-emnlp.134/",
    doi = "10.18653/v1/2023.findings-emnlp.134",
    pages = "2027--2051"
}

@InProceedings{pmlr-v119-hu20b,
  title = 	 {{XTREME}: A Massively Multilingual Multi-task Benchmark for Evaluating Cross-lingual Generalisation},
  author =       {Hu, Junjie and Ruder, Sebastian and Siddhant, Aditya and Neubig, Graham and Firat, Orhan and Johnson, Melvin},
  booktitle = 	 {Proceedings of the 37th International Conference on Machine Learning},
  pages = 	 {4411--4421},
  year = 	 {2020},
  editor = 	 {III, Hal Daumé and Singh, Aarti},
  volume = 	 {119},
  series = 	 {Proceedings of Machine Learning Research},
  month = 	 {13--18 Jul},
  publisher =    {PMLR},
  url = 	 {https://proceedings.mlr.press/v119/hu20b.html}
}

@inproceedings{conneau-etal-2020-unsupervised,
    title = "Unsupervised Cross-lingual Representation Learning at Scale",
    author = "Conneau, Alexis  and
      Khandelwal, Kartikay  and
      Goyal, Naman  and
      Chaudhary, Vishrav  and
      Wenzek, Guillaume  and
      Guzm{\'a}n, Francisco  and
      Grave, Edouard  and
      Ott, Myle  and
      Zettlemoyer, Luke  and
      Stoyanov, Veselin",
    editor = "Jurafsky, Dan  and
      Chai, Joyce  and
      Schluter, Natalie  and
      Tetreault, Joel",
    booktitle = "Proceedings of the 58th Annual Meeting of the Association for Computational Linguistics",
    month = jul,
    year = "2020",
    address = "Online",
    publisher = "Association for Computational Linguistics",
    url = "https://aclanthology.org/2020.acl-main.747/",
    doi = "10.18653/v1/2020.acl-main.747",
    pages = "8440--8451"
}

@inproceedings{ratinov-roth-2009-design,
    title = "Design Challenges and Misconceptions in Named Entity Recognition",
    author = "Ratinov, Lev  and
      Roth, Dan",
    editor = "Stevenson, Suzanne  and
      Carreras, Xavier",
    booktitle = "Proceedings of the Thirteenth Conference on Computational Natural Language Learning ({C}o{NLL}-2009)",
    month = jun,
    year = "2009",
    address = "Boulder, Colorado",
    publisher = "Association for Computational Linguistics",
    url = "https://aclanthology.org/W09-1119/",
    pages = "147--155"
}

@inproceedings{wolf-etal-2020-transformers,
    title = "Transformers: State-of-the-Art Natural Language Processing",
    author = "Wolf, Thomas  and
      Debut, Lysandre  and
      Sanh, Victor  and
      Chaumond, Julien  and
      Delangue, Clement  and
      Moi, Anthony  and
      Cistac, Pierric  and
      Rault, Tim  and
      Louf, Remi  and
      Funtowicz, Morgan  and
      Davison, Joe  and
      Shleifer, Sam  and
      von Platen, Patrick  and
      Ma, Clara  and
      Jernite, Yacine  and
      Plu, Julien  and
      Xu, Canwen  and
      Le Scao, Teven  and
      Gugger, Sylvain  and
      Drame, Mariama  and
      Lhoest, Quentin  and
      Rush, Alexander",
    editor = "Liu, Qun  and
      Schlangen, David",
    booktitle = "Proceedings of the 2020 Conference on Empirical Methods in Natural Language Processing: System Demonstrations",
    month = oct,
    year = "2020",
    address = "Online",
    publisher = "Association for Computational Linguistics",
    url = "https://aclanthology.org/2020.emnlp-demos.6/",
    doi = "10.18653/v1/2020.emnlp-demos.6",
    pages = "38--45"
}

@misc{marone2025mmbertmodernmultilingualencoder,
      title={mmBERT: A Modern Multilingual Encoder with Annealed Language Learning}, 
      author={Marc Marone and Orion Weller and William Fleshman and Eugene Yang and Dawn Lawrie and Benjamin Van Durme},
      year={2025},
      eprint={2509.06888},
      archivePrefix={arXiv},
      primaryClass={cs.CL},
      url={https://arxiv.org/abs/2509.06888}, 
}

@inproceedings{
chung2021rethinking,
title={Rethinking Embedding Coupling in Pre-trained Language Models},
author={Hyung Won Chung and Thibault Fevry and Henry Tsai and Melvin Johnson and Sebastian Ruder},
booktitle={International Conference on Learning Representations},
year={2021},
url={https://openreview.net/forum?id=xpFFI_NtgpW}
}

@inproceedings{tedeschi-etal-2021-wikineural-combined,
    title = "{W}iki{NE}u{R}al: {C}ombined Neural and Knowledge-based Silver Data Creation for Multilingual {NER}",
    author = "Tedeschi, Simone  and
      Maiorca, Valentino  and
      Campolungo, Niccol{\`o}  and
      Cecconi, Francesco  and
      Navigli, Roberto",
    editor = "Moens, Marie-Francine  and
      Huang, Xuanjing  and
      Specia, Lucia  and
      Yih, Scott Wen-tau",
    booktitle = "Findings of the Association for Computational Linguistics: EMNLP 2021",
    month = nov,
    year = "2021",
    address = "Punta Cana, Dominican Republic",
    publisher = "Association for Computational Linguistics",
    url = "https://aclanthology.org/2021.findings-emnlp.215/",
    doi = "10.18653/v1/2021.findings-emnlp.215",
    pages = "2521--2533"
}

@misc{lin2018focallossdenseobject,
      title={Focal Loss for Dense Object Detection}, 
      author={Tsung-Yi Lin and Priya Goyal and Ross Girshick and Kaiming He and Piotr Dollár},
      year={2018},
      eprint={1708.02002},
      archivePrefix={arXiv},
      primaryClass={cs.CV},
      url={https://arxiv.org/abs/1708.02002}, 
}

@inproceedings{qi-etal-2020-stanza,
    title = "{S}tanza: A Python Natural Language Processing Toolkit for Many Human Languages",
    author = "Qi, Peng  and
      Zhang, Yuhao  and
      Zhang, Yuhui  and
      Bolton, Jason  and
      Manning, Christopher D.",
    editor = "Celikyilmaz, Asli  and
      Wen, Tsung-Hsien",
    booktitle = "Proceedings of the 58th Annual Meeting of the Association for Computational Linguistics: System Demonstrations",
    month = jul,
    year = "2020",
    address = "Online",
    publisher = "Association for Computational Linguistics",
    url = "https://aclanthology.org/2020.acl-demos.14/",
    doi = "10.18653/v1/2020.acl-demos.14",
    pages = "101--108"
}

@inproceedings{pytorch,
author = {Ansel, Jason and Yang, Edward and He, Horace and Gimelshein, Natalia and Jain, Animesh and Voznesensky, Michael and Bao, Bin and Bell, Peter and Berard, David and Burovski, Evgeni and Chauhan, Geeta and Chourdia, Anjali and Constable, Will and Desmaison, Alban and DeVito, Zachary and Ellison, Elias and Feng, Will and Gong, Jiong and Gschwind, Michael and Hirsh, Brian and Huang, Sherlock and Kalambarkar, Kshiteej and Kirsch, Laurent and Lazos, Michael and Lezcano, Mario and Liang, Yanbo and Liang, Jason and Lu, Yinghai and Luk, C. K. and Maher, Bert and Pan, Yunjie and Puhrsch, Christian and Reso, Matthias and Saroufim, Mark and Siraichi, Marcos Yukio and Suk, Helen and Zhang, Shunting and Suo, Michael and Tillet, Phil and Zhao, Xu and Wang, Eikan and Zhou, Keren and Zou, Richard and Wang, Xiaodong and Mathews, Ajit and Wen, William and Chanan, Gregory and Wu, Peng and Chintala, Soumith},
title = {PyTorch 2: Faster Machine Learning Through Dynamic Python Bytecode Transformation and Graph Compilation},
year = {2024},
isbn = {9798400703850},
publisher = {Association for Computing Machinery},
address = {New York, NY, USA},
url = {https://doi.org/10.1145/3620665.3640366},
doi = {10.1145/3620665.3640366},
booktitle = {Proceedings of the 29th ACM International Conference on Architectural Support for Programming Languages and Operating Systems, Volume 2},
pages = {929–947},
numpages = {19},
location = {La Jolla, CA, USA},
series = {ASPLOS '24}
}

\appendix


\section{Implementation and Compute}
\label{sec:implementation_appendix}

For all our experiments, we use the transformers library \citep{wolf-etal-2020-transformers} and PyTorch \citep{pytorch}. We conducted all experiments on 8 NVIDIA RTX A6000 GPUs with 48GB VRAM each.

\paragraph{LLM prompting and decoding.}~For all generative baselines, we use each model's default generation hyperparameters as shipped with its chat template, and re-use the extraction prompt and substring-matching parser introduced by \citet{golde2025finerwebdatasetsartifactsscalable}. We prompt the model once per input sentence with the full label set of the respective benchmark and ask it to return the surface forms of all entities per label. The parser then maps each returned surface form back to character offsets in the input by exact substring matching, discarding generations that do not occur verbatim in the input. We use no few-shot examples, self-consistency, or chain-of-thought decoding, so the comparison reflects single-pass prompted extraction. We show the verbatim prompt template in \Cref{fig:llm_prompt}.

\begin{figure}[h]
\centering
\begin{quote}\scriptsize
\begin{verbatim}
<instruction>
Given the following passage in {language}, extract all
named entities and assign each a type. Identify a
diverse and comprehensive set of entities, including
but not limited to:
- Persons, organizations, and locations;
- Technologies, programming languages, and frameworks;
- Scientific concepts, theories, and discoveries;
- Cultural references, works of art, and media;
- Products, brands, and inventions;
- Events, time periods, and dates;
- Quantities, measurements, and statistics;
- Any other meaningful or contextually significant
  named entities.
For each entity, determine the most appropriate and
specific type label. Go beyond general categories when
possible (e.g., use Quantum Theory instead of just
Scientific Concept).
</instruction>
<example>
The output should be a list of tuples in the following
format:
[("entity 1", "type of entity 1"), ...]
</example>
<input>
Passage: {input_passage}
</input>
Annotations:
\end{verbatim}
\end{quote}
\caption{Verbatim extraction prompt used for all generative baselines, following \citet{golde2025finerwebdatasetsartifactsscalable}. The placeholders \texttt{\{language\}} and \texttt{\{input\_passage\}} are filled with the language and the input sentence of the respective benchmark instance.}
\label{fig:llm_prompt}
\end{figure}

\section{Evaluation Benchmarks}~We show an overview of all evaluation benchmarks used in \Cref{tab:ner-datasets}.

\begin{table}[h]
\centering
\begin{tabular}{lcc}
\toprule
\textit{Dataset} & \textit{\# Langs} & \textit{\# Labels} \\
\midrule
PAN-X           & 176 & 3   \\
MasakhaNER      & 20  & 4   \\
UNER            & 13  & 3   \\
MultiCoNER v2   & 12  & 33  \\
MultiCoNER v1   & 11  & 6   \\
MultiNERD       & 10  & 15  \\
DynamicNER      & 8   & 155 \\
\bottomrule
\end{tabular}
\caption{Overview of evaluation benchmarks used in our experiments, showing the number of supported languages and the number of entity types.}
\label{tab:ner-datasets}
\end{table}

\section{Loss Function Definitions} \label{sec:loss_functions_appendix}

For each candidate span $s_{i,j}$, our models predict a score for every entity label $e_k$. We denote by $p_{i,j,k}$ the predicted probability for span $s_{i,j}$ for label $e_k$.
Further, let $k^{*}$ denote the gold label index for span $s_{i,j}$ (or $\varnothing$ if the span is negative).

\paragraph{Binary cross-entropy loss.}
For a given span $s_{i,j}$ with gold label $k^{*}$, binary cross-entropy computes the negative log-likelihood over the model’s predicted scores for each label:
\begin{gather}
\ell_{\text{BCE}}
= - \log p_{i,j,k^{*}}
- \sum_{k \neq k^{*}} \log \bigl(1 - p_{i,j,k}\bigr) .
\end{gather}

\paragraph{Focal loss.}~For a given span $s_{i,j}$ and a label $k$, we define
\begin{gather}
p_t =
\begin{cases}
p_{i,j,k^{*}} & \text{if } k = k^{*},\\
1 - p_{i,j,k} & \text{otherwise}.
\end{cases}
\end{gather}
The focal loss is then
\begin{gather}
\ell_{\text{Focal}}(p_t) = - \alpha_t (1 - p_t)^{\gamma} \log(p_t).
\end{gather}
The focusing parameter $\gamma$ controls the down-weighting of easy examples, and for $\gamma = 0$ the loss reduces to classical cross-entropy. The weighting factor $\alpha$ balances positive and negative examples and can be set using inverse class frequencies or tuned as a hyperparameter. This loss follows the idea of \citet{lin2018focallossdenseobject}.

\paragraph{Contrastive loss.}
Following \citet{zhang2023optimizing}, we treat $(s_{i,j}, e_{k^{*}})$ as a positive pair and contrast it against a set of negative spans $S^{-}_{k^{*}}$ for the same label, where $\mathrm{sim}(\cdot,\cdot) = \cos(\cdot,\cdot)/\tau$ is a temperature-scaled cosine similarity. This yields the span-based typing objective:
\begin{gather}
\ell_{\text{span}}
= - \log
\frac{
\exp\bigl(\mathrm{sim}(s_{i,j}, e_{k^{*}})\bigr)
}{
\sum\limits_{s' \in S^{-}_{k^{*}} \cup \{s_{i,j}\}}
\exp\bigl(\mathrm{sim}(s', e_{k^{*}})\bigr)
} .
\end{gather}
The typing objective pushes entity spans close to their label embedding, but does not specify how close a span must be to be predicted as positive at test time. To separate entity from non-entity spans without an explicit \texttt{Outside} class, we add a dynamic thresholding objective. We use the similarity between a special threshold span $s_{0,0}$ (derived from the \texttt{[CLS]} token, which summarizes the full input) and the label embedding $e_{k^{*}}$ as a per-label, input-specific threshold, and learn it jointly with the typing objective:
\begin{gather}
\ell_{\text{thr}}
= - \log
\frac{
\exp\bigl(\mathrm{sim}(s_{0,0}, e_{k^{*}})\bigr)
}{
\sum\limits_{s' \in S^{-}_{k^{*}}}
\exp\bigl(\mathrm{sim}(s', e_{k^{*}})\bigr)
} .
\end{gather}
The two objectives are combined into a single contrastive loss,
\begin{gather}
\ell_{\text{Con}} = \alpha\,\ell_{\text{span}} + \beta\,\ell_{\text{thr}},
\end{gather}
where $\alpha$ and $\beta$ weight the typing and the thresholding objective, respectively. We set $\alpha = \beta = 0.5$, giving equal weight to both. At inference, a span is predicted as positive for label $k^{*}$ if $\mathrm{sim}(s_{i,j}, e_{k^{*}}) > \mathrm{sim}(s_{0,0}, e_{k^{*}})$, i.e.\ if it is more similar to the label than to the learned threshold.

\paragraph{Dice loss.}
Dice loss optimizes the soft Sørensen–Dice coefficient between predicted and target spans, thus targeting overlap rather than per-pair classification \citep{li-etal-2020-dice}. Let $p_{i,j,k} = \sigma(\mathrm{sim}(s_{i,j}, e_k))$ be the predicted probability that span $s_{i,j}$ has label $k$, and let $y_{i,j,k} \in \{0,1\}$ be the corresponding target. The loss is defined as:
\begin{gather}
\ell_{\text{Dice}}
= 1 - \frac{2 \sum_{i,j,k} p_{i,j,k}\, y_{i,j,k} + \gamma}
{\sum_{i,j,k} p_{i,j,k} + \sum_{i,j,k} y_{i,j,k} + \gamma},
\end{gather}
where $\gamma$ is a smoothing constant that stabilizes training and avoids division by zero. Because the score depends on the relative overlap between positives and predictions rather than on every negative span individually, dice loss is naturally robust to the severe positive–negative imbalance in span-based NER.

\section{The Impact of Word-Segmented Inputs}
\label{sec:pre_tokenized_text_impact_appendix}

We do not use word-segmented text when training our models, although this practice is common in prior work. Word segmentation allows masking irrelevant spans during training, for example when a span covers multiple words and each word consists of several subwords (cf.\ \Cref{fig:subword-prefix-miss}). In such cases, the model does not need to compute loss terms for subword combinations that do not form valid spans. However, this approach requires language-specific word-segmentation models. Maintaining such models for more than 100 languages introduces substantial manual overhead. We therefore omit word segmentation and instead let the model learn span boundaries implicitly. Concretely, we treat all subword combinations up to a maximum span length $l$ as candidate spans.

\begin{figure}[t]
    \centering
    \includegraphics[width=\linewidth]{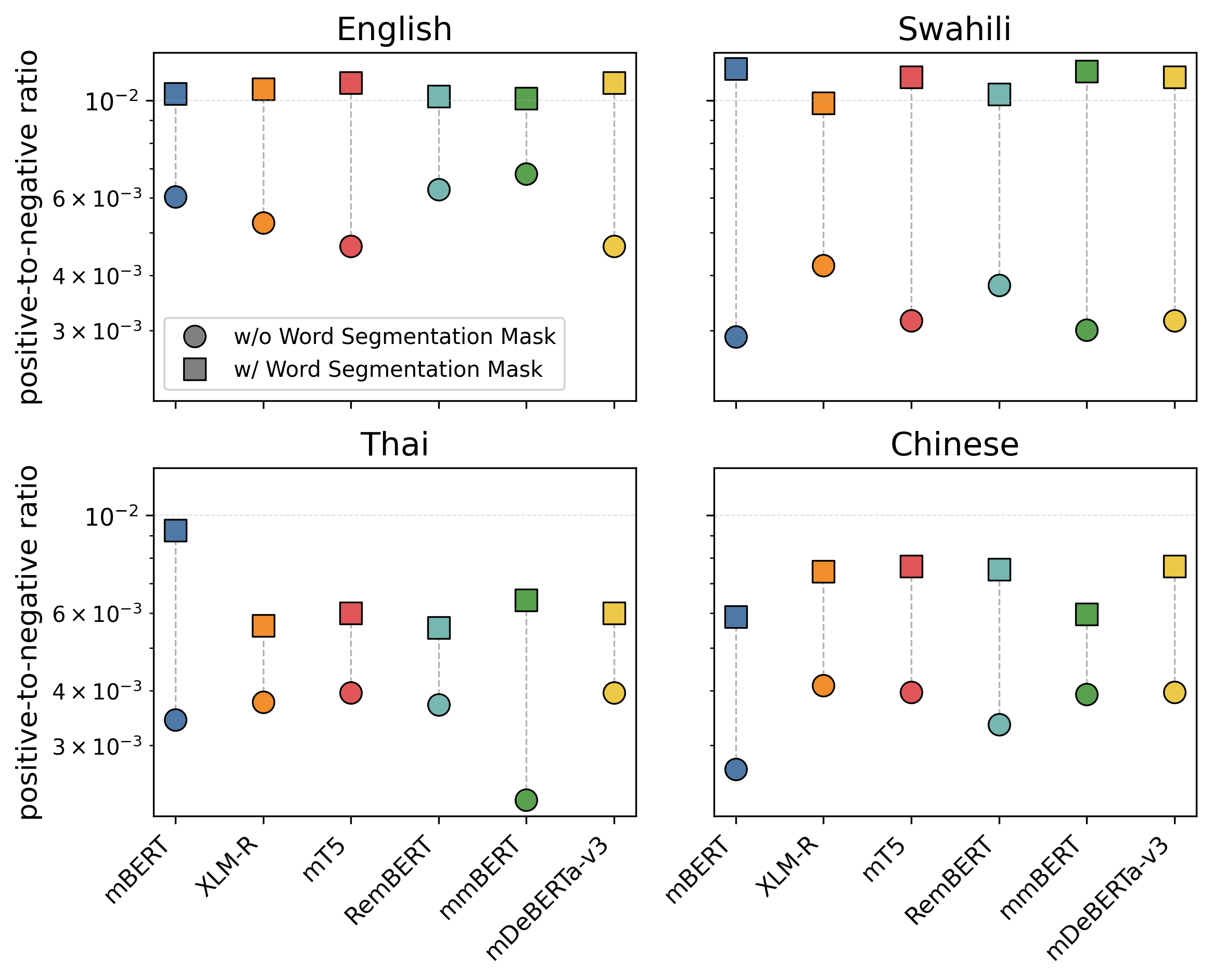}
    \caption{Positive-to-negative span ratios per language and tokenizer, with and without a word-segmentation mask. Segmentation raises the ratio in every case, so training on raw text means training under a harsher class imbalance.}
    \label{fig:positive_negative_ratios_tokenize}
\end{figure}

To quantify what this costs us, we select two whitespace-separated languages from FiNERweb (English and Swahili) and two non-whitespace-separated languages (Thai and Chinese). For each language, we subword-tokenize the data in two ways: using raw text with span labels based on character offsets, and using word-segmented inputs with span labels based on word boundaries. We apply Stanza \citep{qi-etal-2020-stanza} for word segmentation in Chinese and Thai, and simple whitespace splitting for English and Swahili. We then remap span annotations from the character level to the token level and enumerate all valid spans up to length $l = 30$, both with and without word-segmentation constraints.

We report the resulting positive-to-negative span ratios in \Cref{fig:positive_negative_ratios_tokenize}. We observe that word segmentation raises the ratio for every language and tokenizer, i.e.\ training on raw text means training under a harsher class imbalance. This is the trade-off we accept in exchange for not maintaining a segmentation model per language, and \Cref{sec:error_analysis_appendix} shows a concrete case in which it hurts.

\section{Score Distributions and Calibration by Architecture}
\label{sec:calibration_appendix}

In \Cref{sec:dim1}, we observe that the bi-encoder peaks at lower decision thresholds than the cross-encoder, and we attribute this to late versus early fusion of span and label information. Here, we verify this directly. Using the trained bi- and cross-encoder checkpoints (mmBERT, PileNER), we score every candidate span--label pair on the eight holdout validation splits (roughly 3.6M candidate pairs per split), label each pair as positive or negative against the gold entities, and compare the resulting score distributions.

\begin{table}[h]
\centering
\small
\begin{tabular}{lcc}
\toprule
\textit{Metric} & Bi-Enc. & Cross-Enc. \\
\midrule
Mean prob.\ on gold spans & 0.384 & \textbf{0.573} \\
Gold spans with $p > 0.5$ & 40.5\% & \textbf{62.1\%} \\
Mean prob.\ on negative spans & $7.2\mathrm{e}{-4}$ & \textbf{$2.4\mathrm{e}{-4}$} \\
Expected calibration error & $5.3\mathrm{e}{-4}$ & \textbf{$1.5\mathrm{e}{-4}$} \\
Score entropy & $2.5\mathrm{e}{-3}$ & \textbf{$8.7\mathrm{e}{-4}$} \\
\bottomrule
\end{tabular}
\caption{Score distribution and calibration statistics aggregated over the eight holdout splits. Lower is better for negative-span probability, calibration error, and entropy.}
\label{tab:calibration_stats}
\end{table}

We report the resulting statistics in \Cref{tab:calibration_stats}. The cross-encoder assigns gold-positive spans a mean probability of 0.573 against 0.384 for the bi-encoder, and 62.1\% of its true positives exceed 0.5 compared to only 40.5\%. It is also better calibrated (expected calibration error $1.5\mathrm{e}{-4}$ vs.\ $5.3\mathrm{e}{-4}$), sharper (mean binary entropy $8.7\mathrm{e}{-4}$ vs.\ $2.5\mathrm{e}{-3}$), and pushes negatives further down (mean probability $2.4\mathrm{e}{-4}$ vs.\ $7.2\mathrm{e}{-4}$). This explains the threshold behavior observed in \Cref{fig:threshold_comparison}: the correct matches of the bi-encoder cluster below 0.5, so its optimum sits at a lower threshold, whereas those of the cross-encoder cluster near 1.0 and thus favor a higher one.

\section{Subword Fragmentation and Optimal Thresholds}
\label{sec:fragmentation_appendix}

In \Cref{sec:dim1}, we further observe that non-Latin scripts peak at lower thresholds, and we attribute this to subword fragmentation. To verify this, we correlate the per-language subword fragmentation rate, measured as the number of subword tokens produced per character of raw text, with the per-language optimal threshold from \Cref{fig:threshold_per_language_with_pilener}.

\begin{table}[h]
\centering
\small
\begin{tabular}{llcc}
\toprule
& & \textit{Subwords} & \textit{Optimal} \\
\textit{Lang.} & \textit{Script} & \textit{/ char} & \textit{threshold} \\
\midrule
eng & Latin       & 0.263 & 0.60 \\
deu & Latin       & 0.283 & 0.60 \\
spa & Latin       & 0.289 & 0.55 \\
fra & Latin       & 0.338 & 0.45 \\
ara & Arabic      & 0.442 & 0.45 \\
hin & Devanagari  & 0.548 & 0.50 \\
jpn & CJK         & 1.118 & 0.25 \\
zho & CJK         & 1.175 & 0.30 \\
\bottomrule
\end{tabular}
\caption{Subword fragmentation rate and optimal decision threshold per holdout language, sorted by fragmentation.}
\label{tab:fragmentation}
\end{table}

We report both quantities in \Cref{tab:fragmentation} and observe a monotone trend. Chinese and Japanese text is fragmented most heavily (about 1.1 subwords per character) and peaks at the lowest thresholds ($0.25$--$0.30$), while Latin-script text is fragmented least ($0.26$--$0.34$) and peaks highest ($0.45$--$0.60$). Arabic and Devanagari fall in between on both axes. This is the mechanism we appeal to in \Cref{sec:dim1}: more subword fragments per unit of text produce more candidate spans per entity, which spreads probability mass and thus lowers the optimal threshold.

\section{Gradient Updates in Multilingual Models} \label{sec:subword_updates}

We show in \Cref{tab:subword_distribution} the percentage of subword embeddings that receive gradient updates for each of the three fine-tuning datasets.

\begin{table}[t]
\centering
\begin{tabular}{lc}
\toprule
 & \textit{Subword Tokens} \\
 \textit{Dataset} & \textit{With Gradients (\%)} \\
\midrule
PileNER  & 38.55 \\
Euro-GLiNER-x & 46.30 \\
FiNERweb & 94.16 \\
\bottomrule
\end{tabular}
\caption{Fraction of subword embeddings updated during training.}
\label{tab:subword_distribution}
\end{table}

\section{Train-Eval Distributional Similarity}
\label{sec:similarity_appendix}

To verify that the improvements from multilingual training cannot be explained by distributional matching between the training and the evaluation data, we measure the lexical and semantic similarity between the training corpora and the evaluation benchmark. We use the English, German, and Russian evaluation splits of MultiNERD and compare each against same-language samples drawn from PileNER, Euro-GLiNER-x, and FiNERweb, using 20 subsamples of 5{,}000 sentences per language. We report two metrics: \textit{(i)} the character 3--5-gram Jensen--Shannon divergence between the n-gram distributions, capturing lexical similarity, and \textit{(ii)} the Fréchet distance between the sentence-embedding distributions, capturing semantic similarity. We compute sentence embeddings with paraphrase-multilingual-MiniLM-L12-v2. For both metrics, lower values indicate greater similarity to MultiNERD. PileNER is English-only and Euro-GLiNER-x does not cover Russian, so the corresponding cells are empty.

\begin{table}[h]
\centering
\begin{tabular}{lccc}
\toprule
\textit{Training Corpus} & ENG & DEU & RUS \\
\midrule
\multicolumn{4}{l}{\textit{Lexical (n-gram JS divergence)}} \\
FiNERweb & 0.153 & 0.170 & 0.235 \\
Euro-GLiNER-x & 0.157 & 0.136 & -- \\
PileNER & 0.160 & -- & -- \\
\midrule
\multicolumn{4}{l}{\textit{Semantic (Fréchet distance)}} \\
FiNERweb & 2.339 & 1.844 & 2.413 \\
Euro-GLiNER-x & 0.959 & 0.379 & -- \\
PileNER & 3.545 & -- & -- \\
\bottomrule
\end{tabular}
\caption{Lexical and semantic distance between each training corpus and the MultiNERD evaluation splits (English, German, Russian), averaged over 20 subsamples of 5{,}000 sentences per language. Lower is more similar.}
\label{tab:similarity}
\end{table}

We report the results in \Cref{tab:similarity} and find that FiNERweb is not the closest corpus to MultiNERD on either metric. On the semantic measure, Euro-GLiNER-x is substantially closer to MultiNERD than FiNERweb for both English (0.959 vs.\ 2.339) and German (0.379 vs.\ 1.844), and lexically the three corpora are comparable, with FiNERweb only marginally closer for English. Despite this, FiNERweb yields the strongest downstream performance, particularly on non-Latin scripts. We therefore conclude that the gains from broad multilingual training do not stem from FiNERweb being more distributionally similar to the evaluation data, but from genuinely broader language and script coverage.

\section{Candidate-Span Imbalance and the Cross-Encoder Advantage}
\label{sec:negpos_appendix}

In \Cref{sec:combining}, we attribute the advantage of the cross-encoder at scale to its ability to capture the varying positive-to-negative span ratio across languages. To verify this, we compute the per-language negative-to-positive candidate-span ratio on the holdout splits and correlate it with the per-language F1 gap between the cross- and the bi-encoder.

\begin{table}[h]
\centering
\small
\begin{tabular}{llcc}
\toprule
\textit{Lang.} & \textit{Script} & \textit{neg:pos} & $\Delta$F1 \\
\midrule
fra & Latin      & 56  & +0.159 \\
ara & Arabic     & 88  & +0.162 \\
hin & Devanagari & 112 & +0.217 \\
eng & Latin      & 264 & +0.089 \\
deu & Latin      & 276 & +0.121 \\
spa & Latin      & 353 & +0.042 \\
jpn & CJK        & 473 & +0.074 \\
zho & CJK        & 884 & +0.023 \\
\bottomrule
\end{tabular}
\caption{Negative-to-positive candidate-span ratio and cross-encoder minus bi-encoder F1 per holdout language, sorted by ratio.}
\label{tab:negpos}
\end{table}

We report both quantities in \Cref{tab:negpos} and observe that the cross-encoder advantage is largest where positive spans are dense (low negative-to-positive ratio) and smallest where they are sparse, which is consistent with the explanation given in \Cref{sec:combining}. We note one caveat: the low-ratio languages here (French, Arabic, and Hindi) also tend to have shorter texts in PAN-X, so the ratio is partly correlated with sequence length. We therefore report this as supporting rather than as conclusive evidence.

\section{Smaller and Quantized LLM Baselines}
\label{sec:efficiency_appendix}

Smaller and quantized LLMs are an important point of comparison for efficiency, and we already use FP8-quantized versions of Qwen and Gemma in our main experiments. To further isolate the efficiency trade-off, we conduct an additional ablation on the English, German, and Russian splits of MultiNERD (1{,}000 examples each), comparing our best-performing cross-encoder against the smaller Qwen3-0.6B and Qwen3-4B variants. We measure latency per example as a single encoder forward pass for \methodname{} and as a single generation step for the autoregressive models, using batch size 1 for a fair comparison. We take all measurements on a single H100.

\begin{table}[h]
\centering
\small
\begin{tabular}{lccc}
\toprule
\textit{Model} & F1 & Latency (ms) & VRAM (GB) \\
\midrule
\multicolumn{4}{l}{\textit{English}} \\
Qwen3-0.6B & 0.197 & 1029.9 & 1.25 \\
Qwen3-4B & 0.475 & 1111.1 & 7.73 \\
Otter & \textbf{0.769} & \textbf{18.5} & 1.43 \\
\midrule
\multicolumn{4}{l}{\textit{German}} \\
Qwen3-0.6B & 0.241 & 939.6 & 1.20 \\
Qwen3-4B & 0.484 & 1316.2 & 7.75 \\
Otter & \textbf{0.709} & \textbf{18.3} & 1.44 \\
\midrule
\multicolumn{4}{l}{\textit{Russian}} \\
Qwen3-0.6B & 0.043 & 1526.1 & 1.25 \\
Qwen3-4B & 0.274 & 1338.1 & 7.75 \\
Otter & \textbf{0.588} & \textbf{18.3} & 1.45 \\
\bottomrule
\end{tabular}
\caption{F1, per-example latency (batch size 1, single H100), and VRAM on the English, German, and Russian splits of MultiNERD (1{,}000 examples each). Best per language in bold.}
\label{tab:efficiency}
\end{table}

We report the results in \Cref{tab:efficiency}. Even against smaller or quantized LLMs, the autoregressive baselines either incur substantially higher latency or achieve lower performance than the distilled encoder-only model. \methodname{} outperforms Qwen3-4B by a wide margin in F1 on all three languages while being roughly 50--80 times faster per example and using a fraction of the memory. We do not report exact latencies for our largest baselines (about 30B parameters), since we ran these through inference providers. Our local FP8 measurements place their per-example latency on the order of several seconds, although this could be reduced with batching. Overall, these results support our claim that distilled encoder models offer a substantially better performance-to-cost trade-off for large-scale multilingual NER.

\section{Error Analysis}
\label{sec:error_analysis_appendix}
~As we do not rely on word segmentation and instead learn entity boundaries implicitly at the embedding level, we observe negative implications for languages with productive prefixation. For example, on the Shona split of MasakhaNER, we find that our model achieves an F1 score of 0.298, which is substantially lower than comparable baselines such as GLiNER (above 0.6 F1 for both variants). By inspecting the model predictions, we observe that locative or associative prefixes attached to named entities (e.g., \textit{eZimbabwe} “in Zimbabwe”, \textit{paVaMugabe} “at Mr.~Mugabe”) lead to discrepancies between our predicted spans and the evaluation format. Considering the range of languages covered by the training data, we find that such constructions are comparatively rare, which leads the model to predict entity spans starting at the lexical stem (e.g., \textit{Zimbabwe Republic Police}) and thus results in boundary mismatches with the gold annotations (\Cref{fig:subword-prefix-miss}). While this behavior does not conform to the annotation guidelines, we consider it linguistically plausible.

\definecolor{correctORG}{RGB}{210,245,210} 
\definecolor{errorO}{RGB}{255,200,200}     

\newcommand{\ok}[1]{\colorbox{correctORG}{#1}}
\newcommand{\err}[1]{\colorbox{errorO}{#1}}

\begin{figure}[t]
\centering
\begin{quote}\small
\begingroup
\setlength{\parskip}{0pt}\setlength{\parindent}{0pt}

\textbf{Original Inputs:} [``Mapurisa'', ``eZimbabwe'', ``Republic'', ``Police'', \ldots] \\
\textbf{Gold:} \texttt{[O, ORG, ORG, ORG, \ldots]}\\
\textbf{Subword Tokenized:} [\texttt{\_Map}, \texttt{ur}, \texttt{isa}, \texttt{\err{\_e}}, \texttt{Zimbabwe}, \texttt{\_Republic}, \texttt{\_Police}, \ldots]\\
\textbf{Pred:}
[\texttt{O, O, O, \err{O}, \ok{ORG}, \ok{ORG}, \ok{ORG}}]

\endgroup
\end{quote}
\caption{Example of a subword-boundary error. The prefix marker in \texttt{eZimbabwe} is labeled as \textsc{O} (red), while the following subwords are correctly predicted as \textsc{ORG} (green).}
\label{fig:subword-prefix-miss}
\end{figure}

\section{Detailed Results} \label{sec:full_experimental_results_appendix}

We show detailed results for selected experiments in \Cref{tab:detailed_results_mmbert,tab:detailed_results_rembert}.

{\renewcommand{\arraystretch}{1.25}
\begin{table*}
\centering
\small
\begin{tabular}{lll|ccccccc|r}
\toprule
\textsc{Arc}. & \textsc{Data-} & $\tau$ &\textsc{Dynamic-} & \textsc{Masakha-} & \multicolumn{2}{c}{\textsc{MultiCoNER}} & \textsc{Multi-} & \textsc{PAN-X} & \textsc{UNER} & \textsc{Avg.} \\
 & \textsc{set} & &\textsc{NER} & \textsc{NER} & \textsc{v1} & \textsc{v2} & \textsc{NERD} & & & \\
\midrule
\multirow[c]{21}{*}{\rotatebox{90}{{\centering Bi-Encoder}}} & \multirow[c]{7}{*}{\rotatebox{90}{{\centering PileNER}}} & 0.050 & 0.107 & 0.495 & 0.277 & 0.095 & 0.464 & 0.420 & 0.560 & 0.345 \\
 &  & 0.100 & 0.113 & 0.514 & 0.289 & 0.095 & 0.534 & 0.418 & 0.614 & 0.368 \\
 &  & 0.150 & 0.113 & 0.517 & 0.287 & 0.090 & 0.577 & 0.410 & 0.642 & 0.377 \\
 &  & \cellcolor{lightgray}0.200 & \cellcolor{lightgray}0.113 & \cellcolor{lightgray}0.516 & \cellcolor{lightgray}0.279 & \cellcolor{lightgray}0.082 & \cellcolor{lightgray}0.606 & \cellcolor{lightgray}0.402 & \cellcolor{lightgray}0.655 & \cellcolor{lightgray}0.379 \\
 &  & 0.300 & 0.110 & 0.500 & 0.258 & 0.062 & 0.633 & 0.377 & 0.667 & 0.373 \\
 &  & 0.400 & 0.099 & 0.470 & 0.233 & 0.038 & 0.637 & 0.352 & 0.662 & 0.356 \\
 &  & 0.500 & 0.082 & 0.430 & 0.202 & 0.019 & 0.627 & 0.321 & 0.645 & 0.332 \\
\cline{2-11}
 & \multirow[c]{7}{*}{\rotatebox{90}{{\centering Euro-GLiNER-x}}} & 0.050 & 0.064 & 0.549 & 0.261 & 0.100 & 0.675 & 0.490 & 0.700 & 0.406 \\
 &  & 0.100 & 0.059 & 0.552 & 0.263 & 0.101 & 0.699 & 0.488 & 0.718 & 0.411 \\
 &  & \cellcolor{lightgray}0.150 & \cellcolor{lightgray}0.053 & \cellcolor{lightgray}0.553 & \cellcolor{lightgray}0.260 & \cellcolor{lightgray}0.100 & \cellcolor{lightgray}0.710 & \cellcolor{lightgray}0.484 & \cellcolor{lightgray}0.724 & \cellcolor{lightgray}0.412 \\
 &  & 0.200 & 0.046 & 0.550 & 0.253 & 0.097 & 0.718 & 0.480 & 0.728 & 0.410 \\
 &  & 0.300 & 0.042 & 0.544 & 0.230 & 0.091 & 0.730 & 0.470 & 0.729 & 0.405 \\
 &  & 0.400 & 0.036 & 0.537 & 0.212 & 0.086 & 0.740 & 0.461 & 0.731 & 0.400 \\
 &  & 0.500 & 0.034 & 0.529 & 0.198 & 0.079 & 0.750 & 0.449 & 0.735 & 0.396 \\
\cline{2-11}
 & \multirow[c]{7}{*}{\rotatebox{90}{{\centering FiNERweb}}} & 0.050 & 0.145 & 0.505 & 0.328 & 0.137 & 0.491 & 0.510 & 0.608 & 0.389 \\
 &  & 0.100 & 0.158 & 0.538 & 0.346 & 0.149 & 0.543 & 0.517 & 0.677 & 0.418 \\
 &  & 0.150 & 0.164 & 0.545 & 0.352 & 0.154 & 0.574 & 0.515 & 0.702 & 0.430 \\
 &  & \cellcolor{lightgray}0.200 & \cellcolor{lightgray}0.166 & \cellcolor{lightgray}0.541 & \cellcolor{lightgray}0.351 & \cellcolor{lightgray}0.156 & \cellcolor{lightgray}0.595 & \cellcolor{lightgray}0.508 & \cellcolor{lightgray}0.704 & \cellcolor{lightgray}0.432 \\
 &  & 0.300 & 0.164 & 0.516 & 0.331 & 0.144 & 0.612 & 0.489 & 0.681 & 0.420 \\
 &  & 0.400 & 0.160 & 0.468 & 0.287 & 0.113 & 0.603 & 0.451 & 0.632 & 0.388 \\
 &  & 0.500 & 0.145 & 0.397 & 0.233 & 0.067 & 0.557 & 0.394 & 0.541 & 0.333 \\
\cline{1-11} \cline{2-11}
\multirow[c]{21}{*}{\rotatebox{90}{{\centering Cross-Encoder}}} & \multirow[c]{7}{*}{\rotatebox{90}{{\centering PileNER}}} & 0.050 & 0.199 & 0.049 & 0.252 & 0.190 & 0.506 & 0.094 & 0.045 & 0.191 \\
 &  & 0.100 & 0.216 & 0.049 & 0.223 & 0.206 & 0.562 & 0.102 & 0.060 & 0.203 \\
 &  & \cellcolor{lightgray}0.150 & \cellcolor{lightgray}0.229 & \cellcolor{lightgray}0.047 & \cellcolor{lightgray}0.192 & \cellcolor{lightgray}0.209 & \cellcolor{lightgray}0.593 & \cellcolor{lightgray}0.100 & \cellcolor{lightgray}0.069 & \cellcolor{lightgray}0.206 \\
 &  & 0.200 & 0.230 & 0.044 & 0.161 & 0.207 & 0.612 & 0.098 & 0.071 & 0.203 \\
 &  & 0.300 & 0.225 & 0.034 & 0.108 & 0.192 & 0.627 & 0.086 & 0.073 & 0.192 \\
 &  & 0.400 & 0.204 & 0.024 & 0.067 & 0.166 & 0.625 & 0.064 & 0.049 & 0.171 \\
 &  & 0.500 & 0.166 & 0.014 & 0.036 & 0.133 & 0.603 & 0.044 & 0.027 & 0.146 \\
\cline{2-11}
 & \multirow[c]{7}{*}{\rotatebox{90}{{\centering Euro-GLiNER-x}}} & 0.050 & 0.086 & 0.453 & 0.226 & 0.097 & 0.674 & 0.466 & 0.641 & 0.378 \\
 &  & \cellcolor{lightgray}0.100 & \cellcolor{lightgray}0.079 & \cellcolor{lightgray}0.437 & \cellcolor{lightgray}0.177 & \cellcolor{lightgray}0.066 & \cellcolor{lightgray}0.682 & \cellcolor{lightgray}0.444 & \cellcolor{lightgray}0.641 & \cellcolor{lightgray}0.361 \\
 &  & 0.150 & 0.068 & 0.417 & 0.137 & 0.040 & 0.677 & 0.422 & 0.622 & 0.341 \\
 &  & 0.200 & 0.052 & 0.392 & 0.098 & 0.023 & 0.662 & 0.399 & 0.604 & 0.319 \\
 &  & 0.300 & 0.032 & 0.337 & 0.038 & 0.006 & 0.608 & 0.343 & 0.546 & 0.273 \\
 &  & 0.400 & 0.016 & 0.272 & 0.011 & 0.001 & 0.524 & 0.282 & 0.468 & 0.225 \\
 &  & 0.500 & 0.007 & 0.197 & 0.004 & 0.000 & 0.413 & 0.216 & 0.370 & 0.173 \\
\cline{2-11}
 & \multirow[c]{7}{*}{\rotatebox{90}{{\centering FiNERweb}}} & 0.050 & 0.303 & 0.517 & 0.321 & 0.260 & 0.550 & 0.471 & 0.493 & 0.416 \\
 &  & 0.100 & 0.324 & 0.551 & 0.340 & 0.276 & 0.584 & 0.473 & 0.524 & 0.439 \\
 &  & 0.150 & 0.337 & 0.564 & 0.349 & 0.284 & 0.604 & 0.467 & 0.533 & 0.448 \\
 &  & \cellcolor{lightgray}0.200 & \cellcolor{lightgray}0.345 & \cellcolor{lightgray}0.571 & \cellcolor{lightgray}0.351 & \cellcolor{lightgray}0.290 & \cellcolor{lightgray}0.619 & \cellcolor{lightgray}0.460 & \cellcolor{lightgray}0.542 & \cellcolor{lightgray}0.454 \\
 &  & \cellcolor{lightgray}0.300 & \cellcolor{lightgray}0.354 & \cellcolor{lightgray}0.573 & \cellcolor{lightgray}0.347 & \cellcolor{lightgray}0.294 & \cellcolor{lightgray}0.636 & \cellcolor{lightgray}0.436 & \cellcolor{lightgray}0.540 & \cellcolor{lightgray}0.454 \\
 &  & 0.400 & 0.361 & 0.563 & 0.334 & 0.290 & 0.649 & 0.405 & 0.515 & 0.446 \\
 &  & 0.500 & 0.358 & 0.532 & 0.314 & 0.283 & 0.659 & 0.363 & 0.461 & 0.424 \\
\bottomrule
\end{tabular}
\caption{Detailed results for \texttt{rembert-base}.}
\label{tab:detailed_results_rembert}
\end{table*}

{\renewcommand{\arraystretch}{1.25}
\begin{table*}
\centering
\small
\begin{tabular}{lll|ccccccc|r}
\toprule
\textsc{Arc}. & \textsc{Data-} & $\tau$ &\textsc{Dynamic-} & \textsc{Masakha-} & \multicolumn{2}{c}{\textsc{MultiCoNER}} & \textsc{Multi-} & \textsc{PAN-X} & \textsc{UNER} & \textsc{Avg.} \\
 & \textsc{set} & &\textsc{NER} & \textsc{NER} & \textsc{v1} & \textsc{v2} & \textsc{NERD} & & & \\
\midrule
\multirow[c]{21}{*}{\rotatebox{90}{{\centering Bi-Encoder}}} & \multirow[c]{7}{*}{\rotatebox{90}{{\centering PileNER}}} & 0.050 & 0.150 & 0.427 & 0.289 & 0.129 & 0.497 & 0.451 & 0.508 & 0.350 \\
 &  & 0.100 & 0.163 & 0.439 & 0.291 & 0.135 & 0.543 & 0.457 & 0.551 & 0.368 \\
 &  & \cellcolor{lightgray}0.150 & \cellcolor{lightgray}0.169 & \cellcolor{lightgray}0.434 & \cellcolor{lightgray}0.282 & \cellcolor{lightgray}0.136 & \cellcolor{lightgray}0.566 & \cellcolor{lightgray}0.452 & \cellcolor{lightgray}0.574 & \cellcolor{lightgray}0.373 \\
 &  & \cellcolor{lightgray}0.200 & \cellcolor{lightgray}0.170 & \cellcolor{lightgray}0.425 & \cellcolor{lightgray}0.270 & \cellcolor{lightgray}0.134 & \cellcolor{lightgray}0.580 & \cellcolor{lightgray}0.443 & \cellcolor{lightgray}0.592 & \cellcolor{lightgray}0.373 \\
 &  & 0.300 & 0.168 & 0.399 & 0.236 & 0.125 & 0.586 & 0.421 & 0.618 & 0.365 \\
 &  & 0.400 & 0.151 & 0.362 & 0.203 & 0.116 & 0.567 & 0.389 & 0.606 & 0.342 \\
 &  & 0.500 & 0.127 & 0.322 & 0.172 & 0.102 & 0.528 & 0.347 & 0.575 & 0.310 \\
\cline{2-11}
 & \multirow[c]{7}{*}{\rotatebox{90}{{\centering Euro-GLiNER-x}}} & \cellcolor{lightgray}0.050 & \cellcolor{lightgray}0.074 & \cellcolor{lightgray}0.505 & \cellcolor{lightgray}0.298 & \cellcolor{lightgray}0.087 & \cellcolor{lightgray}0.660 & \cellcolor{lightgray}0.496 & \cellcolor{lightgray}0.753 & \cellcolor{lightgray}0.410 \\
 &  & 0.100 & 0.057 & 0.502 & 0.268 & 0.076 & 0.680 & 0.481 & 0.765 & 0.404 \\
 &  & 0.150 & 0.040 & 0.497 & 0.233 & 0.064 & 0.690 & 0.462 & 0.762 & 0.393 \\
 &  & 0.200 & 0.031 & 0.490 & 0.199 & 0.052 & 0.694 & 0.444 & 0.755 & 0.381 \\
 &  & 0.300 & 0.021 & 0.473 & 0.141 & 0.033 & 0.693 & 0.409 & 0.743 & 0.359 \\
 &  & 0.400 & 0.014 & 0.448 & 0.096 & 0.022 & 0.684 & 0.373 & 0.715 & 0.336 \\
 &  & 0.500 & 0.008 & 0.417 & 0.061 & 0.014 & 0.668 & 0.332 & 0.667 & 0.310 \\
\cline{2-11}
 & \multirow[c]{7}{*}{\rotatebox{90}{{\centering FiNERweb}}} & 0.050 & 0.218 & 0.548 & 0.325 & 0.165 & 0.532 & 0.509 & 0.627 & 0.418 \\
 &  & \cellcolor{lightgray}0.100 & \cellcolor{lightgray}0.237 & \cellcolor{lightgray}0.553 & \cellcolor{lightgray}0.343 & \cellcolor{lightgray}0.175 & \cellcolor{lightgray}0.569 & \cellcolor{lightgray}0.509 & \cellcolor{lightgray}0.670 & \cellcolor{lightgray}0.437 \\
 &  & 0.150 & 0.245 & 0.523 & 0.353 & 0.180 & 0.581 & 0.497 & 0.675 & 0.436 \\
 &  & 0.200 & 0.251 & 0.485 & 0.355 & 0.182 & 0.576 & 0.479 & 0.661 & 0.427 \\
 &  & 0.300 & 0.253 & 0.401 & 0.341 & 0.182 & 0.554 & 0.436 & 0.597 & 0.395 \\
 &  & 0.400 & 0.249 & 0.334 & 0.315 & 0.175 & 0.524 & 0.390 & 0.518 & 0.358 \\
 &  & 0.500 & 0.227 & 0.273 & 0.275 & 0.158 & 0.483 & 0.335 & 0.418 & 0.310 \\
\cline{1-11} \cline{2-11}
\multirow[c]{21}{*}{\rotatebox{90}{{\centering Cross-Encoder}}} & \multirow[c]{7}{*}{\rotatebox{90}{{\centering PileNER}}} & 0.050 & 0.223 & 0.321 & 0.175 & 0.084 & 0.477 & 0.362 & 0.468 & 0.301 \\
 &  & \cellcolor{lightgray}0.100 & \cellcolor{lightgray}0.208 & \cellcolor{lightgray}0.359 & \cellcolor{lightgray}0.181 & \cellcolor{lightgray}0.087 & \cellcolor{lightgray}0.536 & \cellcolor{lightgray}0.356 & \cellcolor{lightgray}0.535 & \cellcolor{lightgray}0.323 \\
 &  & 0.150 & 0.174 & 0.351 & 0.175 & 0.082 & 0.562 & 0.338 & 0.545 & 0.318 \\
 &  & 0.200 & 0.137 & 0.327 & 0.162 & 0.071 & 0.572 & 0.314 & 0.534 & 0.303 \\
 &  & 0.300 & 0.080 & 0.265 & 0.131 & 0.049 & 0.571 & 0.263 & 0.473 & 0.262 \\
 &  & 0.400 & 0.036 & 0.195 & 0.096 & 0.029 & 0.547 & 0.216 & 0.370 & 0.213 \\
 &  & 0.500 & 0.015 & 0.125 & 0.062 & 0.015 & 0.503 & 0.167 & 0.244 & 0.162 \\
\cline{2-11}
 & \multirow[c]{7}{*}{\rotatebox{90}{{\centering Euro-GLiNER-x}}} & 0.050 & 0.189 & 0.477 & 0.279 & 0.141 & 0.696 & 0.465 & 0.681 & 0.418 \\
 &  & 0.100 & 0.193 & 0.495 & 0.304 & 0.144 & 0.720 & 0.469 & 0.712 & 0.434 \\
 &  & \cellcolor{lightgray}0.150 & \cellcolor{lightgray}0.193 & \cellcolor{lightgray}0.498 & \cellcolor{lightgray}0.311 & \cellcolor{lightgray}0.138 & \cellcolor{lightgray}0.731 & \cellcolor{lightgray}0.469 & \cellcolor{lightgray}0.727 & \cellcolor{lightgray}0.438 \\
 &  & 0.200 & 0.186 & 0.497 & 0.309 & 0.129 & 0.739 & 0.470 & 0.732 & 0.437 \\
 &  & 0.300 & 0.166 & 0.492 & 0.297 & 0.110 & 0.749 & 0.469 & 0.741 & 0.432 \\
 &  & 0.400 & 0.145 & 0.487 & 0.279 & 0.091 & 0.754 & 0.468 & 0.748 & 0.425 \\
 &  & 0.500 & 0.124 & 0.480 & 0.258 & 0.074 & 0.754 & 0.465 & 0.745 & 0.414 \\
\cline{2-11}
 & \multirow[c]{7}{*}{\rotatebox{90}{{\centering FiNERweb}}} & 0.050 & 0.269 & 0.516 & 0.311 & 0.216 & 0.541 & 0.509 & 0.573 & 0.419 \\
 &  & 0.100 & 0.288 & 0.534 & 0.327 & 0.230 & 0.579 & 0.515 & 0.619 & 0.442 \\
 &  & 0.150 & 0.305 & 0.532 & 0.339 & 0.239 & 0.598 & 0.518 & 0.644 & 0.453 \\
 &  & 0.200 & 0.314 & 0.527 & 0.341 & 0.242 & 0.611 & 0.520 & 0.663 & 0.460 \\
 &  & \cellcolor{lightgray}0.300 & \cellcolor{lightgray}0.325 & \cellcolor{lightgray}0.511 & \cellcolor{lightgray}0.338 & \cellcolor{lightgray}0.233 & \cellcolor{lightgray}0.627 & \cellcolor{lightgray}0.516 & \cellcolor{lightgray}0.678 & \cellcolor{lightgray}0.461 \\
 &  & 0.400 & 0.315 & 0.483 & 0.317 & 0.209 & 0.629 & 0.503 & 0.671 & 0.447 \\
 &  & 0.500 & 0.293 & 0.439 & 0.282 & 0.174 & 0.620 & 0.481 & 0.645 & 0.419 \\
\bottomrule
\end{tabular}
\caption{Detailed results for \texttt{mmBERT-base}.}
\label{tab:detailed_results_mmbert}
\end{table*}

\end{document}